# Rethinking Multimodality: Optimizing Multimodal Deep Learning for Biomedical Signal Classification

[1]Timothy Oladunni, [2]Alex Wong
[1]Computer Science Department, Morgan State University and Yale University
 Timothy.oladunni@morgan.edu; timothy.oladunni@yale.edu

[2]Computer Science Department, Yale University
 alex.wong@yale.edu

**ABSTRACT** This study proposes a novel perspective on multimodal deep learning for biomedical signal classification, systematically analyzing how complementary feature domains impact model performance. While fusing multiple domains often presumes enhanced accuracy, this work demonstrates that adding modalities can yield diminishing returns, as not all fusions are inherently advantageous. To validate this, five deep learning models were designed, developed, and rigorously evaluated: three unimodal (1D-CNN for time, 2D-CNN for time-frequency, and 1D-CNN-Transformer for frequency) and two multimodal (Hybrid 1, which fuses 1D-CNN and 2D-CNN; Hybrid 2, which combines 1D-CNN, 2D-CNN, and a Transformer). For ECG classification, bootstrapping and Bayesian inference revealed that Hybrid 1 consistently outperformed the 2D-CNN baseline across all metrics (p-values < 0.05, Bayesian probabilities > 0.90), confirming the synergistic complementarity of the time and time-frequency domains. Conversely, Hybrid 2's inclusion of the frequency domain offered no further improvement and sometimes a marginal decline, indicating representational redundancy; a phenomenon further substantiated by a targeted ablation study. This research redefines a fundamental principle of multimodal design in biomedical signal analysis. We demonstrate that **optimal domain fusion isn't about the number of modalities, but the quality of their inherent complementarity**. This **paradigm-shifting concept** moves beyond purely heuristic feature selection. Our novel theoretical contribution, "Complementary Feature Domains in Multimodal ECG Deep Learning," presents a mathematically quantifiable framework for identifying ideal domain combinations, demonstrating that optimal multimodal performance arises **from the intrinsic information-theoretic complementarity** among fused domains. The broader applicability of this theory is discussed, extending its relevance beyond ECG to other biomedical domains involving multimodal time-series data.

## 1. INTRODUCTION

Multimodal deep-learning models combine multiple feature domains to build a robust, accurate, and reliable model [1]. This state-of-the-art algorithm has emerged as a promising approach for complex classification tasks, particularly biomedical signal analysis and image recognition [2]. For example, an intermediate fusion approach could be used to build a robust and accurate ECG signal classifier. Domain-specialized neural networks such as 1D-CNN, LSTM, and 2D-CNN are trained to capture features from the time, frequency, and time-frequency domains of the ECG signals, respectively. Features from these three architectures can be concatenated (or fused) to form a larger feature set, with the intuition that concatenated features provide the best representation of the dataset. Models built with concatenated features benefit from the combined strengths of these various data perspectives to create a complete feature set, leading to improved classification accuracy [3]. However, increasing the architectural complexity by adding more domain feature representations may not always translate into better performance. This can result in redundancy, overfitting, and longer training times [4].

The relationship between model complexity and performance has been extensively studied using traditional learning algorithms [5]. However, this is a critical yet underexplored topic in hybrid ECG multimodal deep-learning algorithms. Generally, simpler models are computationally efficient [6] and less prone to overfitting, but are incapable of fully capturing the underlying patterns of complex datasets. However, a hybrid multimodal model is highly complex [7] and capable of leveraging a broader range of features [8], thus risking overfitting to noise, becoming less interpretable, and requiring substantial computational resources [9]. Therefore, there is a need to study the efficient use of these state-of-the-art algorithms.

Goal

While multimodal deep learning has been extensively explored for enhancing ECG classification, a notable gap remains in the literature regarding frameworks for systematically evaluating the trade-offs between model complexity and performance. Previous studies [10, 11] have primarily focused on achieving higher predictive performance in multimodal deep-learning models for ECG signals, without a detailed examination of the efficiency gains or diminishing returns associated with increased architectural complexity or additional modalities. This study directly addresses this gap by establishing a theoretical and empirical framework to assess these critical trade-offs. We systematically investigate whether simply incorporating more domains consistently translates to improved performance.

This work provides a rigorous evaluation of trade-offs in ECG multimodal deep learning models by meticulously examining the impact of various feature domain combinations and model configurations on classification performance. Building on earlier work [12], this analysis explores the complementarity of fused modalities, challenging the common assumption that 'more is better' when evaluating the effectiveness of model fusion. The overarching goal is to offer concrete guidelines for designing efficient and effective hybrid deep learning architectures, thereby advancing their practical application in real-world biomedical scenarios.

Contribution

This work has the following contributions:
- The research 1) utilized 1D-CNN, 2D-CNN, and 1D-CNN-Transformer algorithms to capture feature representations from time, frequency, and time-frequency domains, respectively; 2) designed two robust hybrid multimodal models to integrate diverse data perspectives that yield improved classification results; and 3) identified the conditions under which added model complexity benefits performance and when it leads to diminishing returns.
- The hypothesis of *Complementary Feature Domains for Optimal Multimodal Deep Learning Performance* was empirically investigated using ECG datasets. The experimental findings were validated through statistical analysis employing 1) correlation and mutual information analysis, 2) bootstrapping inference, and 3) Bayesian confidence intervals. An ablation study was conducted to corroborate the results of the statistical significance test, which was subsequently justified by scientific reasoning. Therefore, this study provides an empirical and theoretical framework for ECG multimodal deep-learning model design.
- An ECG signal dataset was used to strategically demonstrate the effectiveness of the proposed methodology in detecting cardiovascular diseases. However, its empirical evidence, statistical analysis, ablation studies, and scientific reasoning render it applicable to other fields that use multidomain hybrid models.
- This study will help researchers and practitioners to determine whether adding feature domains is complementary or redundant. Ultimately, we reinvigorate the discussion on the premise that the optimal performance of a multimodal deep learning model is a function of the complementarity of its fused domains.

This paper is organized as follows. Section 2 discusses the study's methodology, and Sections 3 and 4 highlight discussion and conclusion, respectively.

## 2. METHODOLOGY

### 2.1. Overview

This study systematically evaluates the impact of varying architectural structures on ECG multimodal deep learning. Various complexities emerge from different strategies for feature representation and fusion, which influence a model's ability to learn discriminative patterns. Complementarity or redundancy of concatenated feature domains in a multimodal deep-learning architecture was observed. Feature extraction was performed using signal processing techniques in the time, frequency, and time-frequency domains. This ensures the extraction of relevant signal characteristics from the three different modalities to enhance the classification accuracy and robustness. The temporal and periodic signal components were captured at time and frequency, respectively. Changes in frequency over time were captured in the time-frequency domain [13].

For deep feature extraction [14], specialized architectures include 1D-CNNs for time-domain features, 2D-CNNs for time-frequency representations, and 1D-CNN transformers to capture long-range dependencies. Two hybrid models were designed, developed, and evaluated using different combinations of the feature domain representations.

Each model was trained using the same preprocessed dataset for fair comparison. Trade-offs between added complexity and performance gains were observed and any points of diminishing returns were highlighted. The performance evaluation of the models was based on accuracy, precision, recall, and F1 score. The results were analyzed to identify which combinations of feature domains were complementary, as indicated by significant improvements. Conversely, redundant features were observed, leading to a drop in performance or overfitting. Specifically, complementarity and redundancy in the feature domains were analyzed. This study is in alignment with Occam's razor rule [15].

The experimental design of this study is illustrated in Figure 2. As shown in the diagram, the experiment comprises five steps: a) data acquisition and preprocessing, b) feature extraction from complementary domains, c) feature-level fusion (intermediate fusion), d) classification, and e) model evaluation and performance analysis.

## 2.2. Dataset
The dataset used for this study was obtained from Mendeley Data [16]. The images represent digitally recorded ECG waveforms, captured directly from clinical ECG monitoring systems. Each image preserves key diagnostic signal features, including P-waves, QRS complexes, and T-waves, making them suitable for machine learning-based cardiovascular classification tasks.

The ECG classes considered in this study were as follows:
- ECG Images of Myocardial Infarction Patients:(240x12=2880) – Myocardial Infarction (MI)
- ECG Images of Patients with Abnormal Heartbeat: (233 × 12 =2796) irregular heart rhythms, such as arrhythmias.
- ECG Images of Patients with a History of Myocardial Infarction (MI): (172x12=2064) – Past History of MI
- Normal Person ECG Images: (284x12=3408) – Normal individuals.

Samples of the ECG dataset are shown in Figure 1.

For robust model training and evaluation, data splitting was performed at the patient level. This ensured that all data instances (e.g., multiple images or time points) belonging to a single, unique, pseudonymized patient identifier were allocated exclusively to one of the data partitions. Specifically, the dataset was split into 80% training, 10% validation, and 10% test sets, respectively. Splitting was based on unique patient IDs.

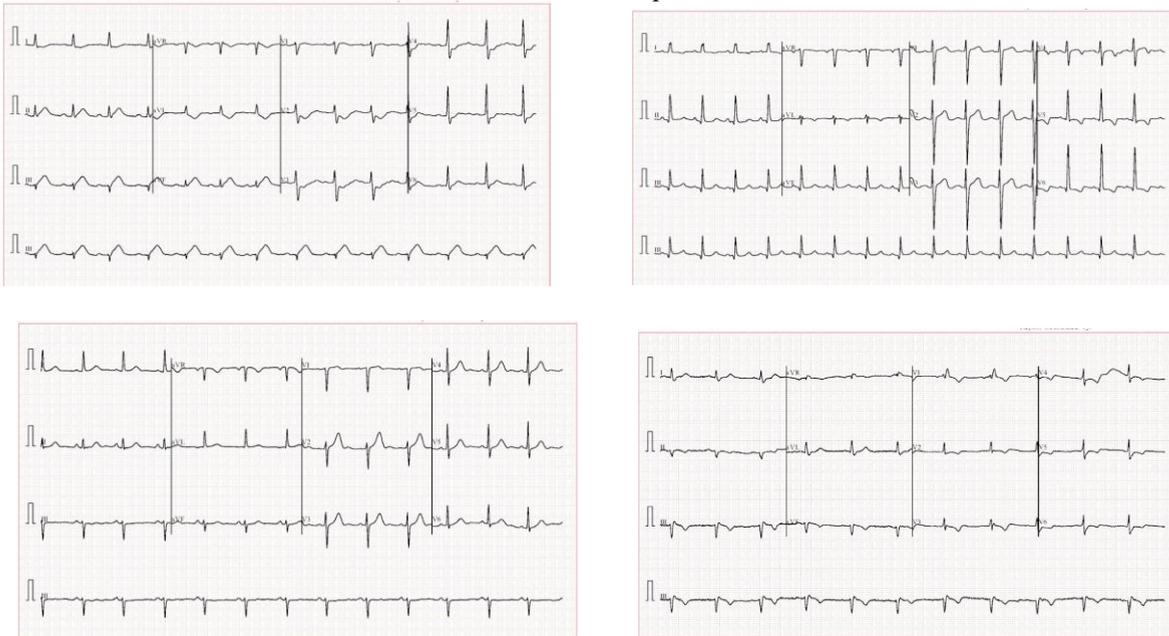

*Figure 1 Samples of the Dataset*

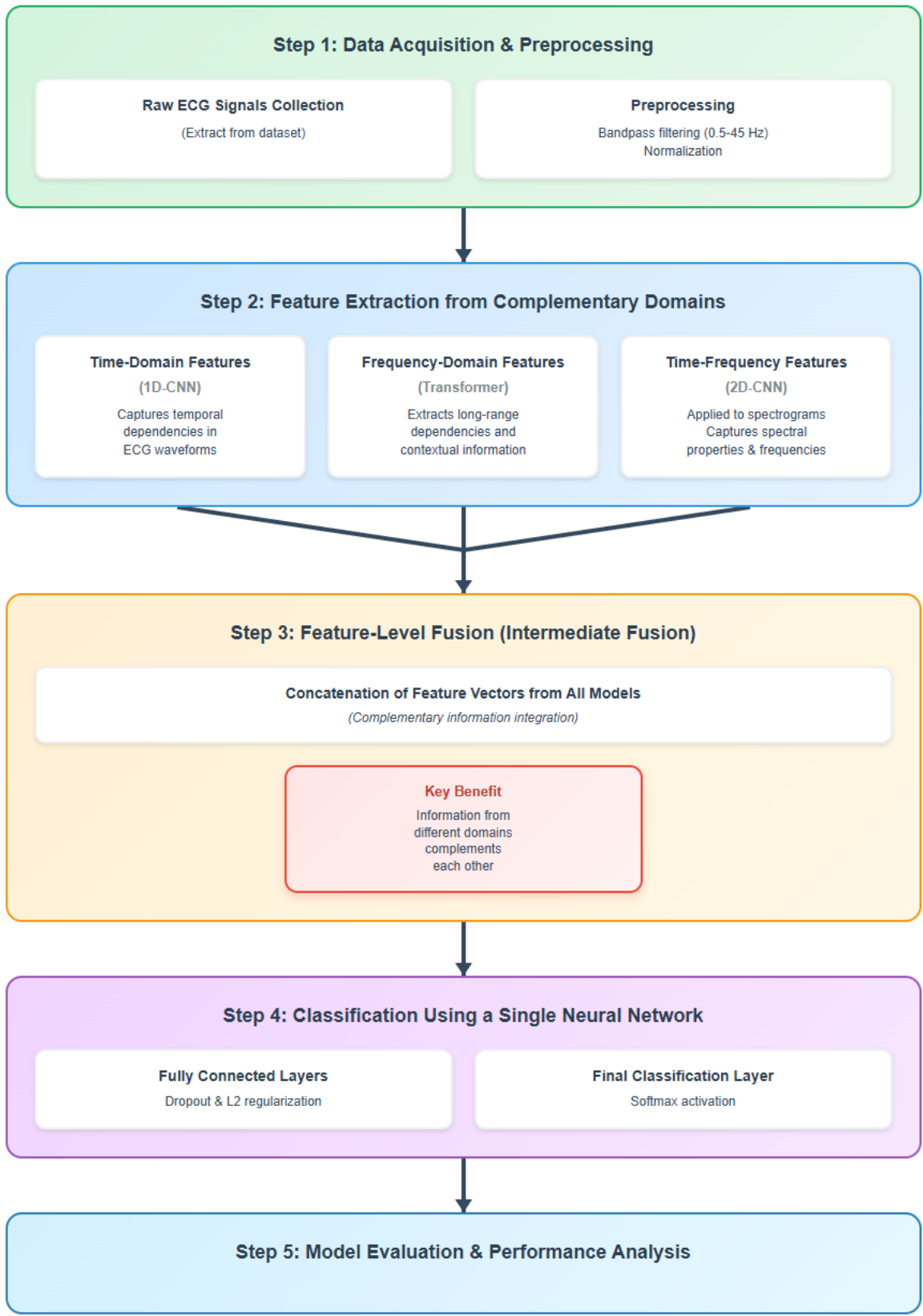

*Figure 2. Experimental Design of the Study*

## 2.3. Dataset Organization and Loading for Experimentation

For experimentation purposes, the dataset was downloaded and organized into four folders, each representing a distinct class ( " Myocardial Infarction (MI), abnormal heartbeat, history of Myocardial Infarction (MI)," and a Normal Person "). Each class label is derived from the name of the folder containing the class. These folders were wrapped in a larger folder named "Dataset" and saved. The hierarchical structure and arrangement facilitate clear class separation and efficient data management and processing. After mounting Google Drive in the Collaboratory environment using the drive.mount ('/content/drive') command, the dataset was loaded into the Colab notebook. Scripting was performed using Python in the Keras/TensorFlow environment. Table 1 details the file paths, loading functions (e.g., (`os.listdir()`, Looping (`for loop`), Conditional Check (`os.path. dir ()`), etc.), and other necessary data preprocessing steps.

*Table 1. Data Loading into Google Collab Environment*

| Task | Description | Python/keras |
|---|---|---|
| **Retrieve Class Labels** | Get the list of class folders in `image_folder`, sorting them for consistency. | File system traversal (`os.listdir()`) |
| **Iterate Over Class Folders** | Loop through each folder, representing a class, to access its images. | Looping (`for loop`), Conditional Check (`os.path.isdir()`) |
| **Read Image Files** | Identify files inside each class folder, construct the file path, and read images. | File Handling (`os.path.join()`), Image Processing (`cv2.imread()`) |
| **Resize Images** | Resize each image to a standard shape (`target_length × target_length`) for uniformity. | Image Scaling (`cv2.resize()`) |

## 2.4. Class Imbalance

A well-documented challenge in machine learning, particularly with medical datasets such as those derived from ECG signals, is the issue of class imbalance [17]. This occurs when certain conditions (e.g., MI, History of MI, Abnormal heartbeat) are significantly rarer than the 'Normal' state. Such an imbalance can have a far-reaching and detrimental effect during model training [18], often leading to models that exhibit high overall accuracy but perform poorly on the critical minority classes due to a bias towards the overwhelming majority [19]. In medical diagnosis, misclassifying a minority class (e.g., a subtle cardiac anomaly) can have severe clinical consequences.

To mitigate this, ADASYN (Adaptive Synthetic Sampling) technique was employed. ADASYN is an advanced oversampling method designed to create synthetic data points for minority classes. ADASYN is particularly well-suited for ECG and other medical time-series data [20] for several key reasons: 1) adaptive sample generation, focusing on "hard" examples, 2) reduced bias and improved generalization, and 3) better handling of complex data distributions. Unlike simpler methods such as SMOTE (Synthetic Minority Over-sampling Technique), which generates synthetic samples uniformly between nearest neighbors, ADASYN adaptively generates *more* synthetic data for minority class samples that are harder to learn [21]. These "hard" examples are typically those lying near the decision boundary or in sparse regions of the feature space. For complex medical signals, such as ECGs, where subtle variations can distinguish between conditions, focusing on these ambiguous or challenging cases enables the model to learn a more robust decision boundary. This is crucial for improving the model's ability to discriminate between clinically significant conditions that might otherwise be overlooked.

In this study, the k parameter was set to 5. This defines the number of nearest neighbors considered when synthesizing new data points. A crucial factor is how ADASYN interprets the local data structure and generates samples. Through this process, the following under-represented medical conditions were augmented:

- Class 0 (MI - Myocardial Infarction): An additional 54 synthetic samples were generated and added to this class.
- Class 1 (History of MI - History of Myocardial Infarction): This class received an augmentation of 56 synthetic samples.
- Class 2 (Abnormal heartbeat): 55 synthetic samples were generated for this class.
- Crucially, Class 3 (Normal), which represented the healthy control group, was already sufficiently populated and well-balanced within the original dataset, thus requiring no synthetic augmentation.

## 2.5. ADASYN Physiological Plausibility

The effectiveness of ADASYN was evaluated using both statistical and visual evidence, as well as the performance of the downstream model.

### 2.5.1 Intra-Class Variance Comparison

Intra-class variance, a fundamental statistical measure of data spread within a defined group [22], was computed for ECG features. The statistical consistency of the generated synthetic samples is assessed by comparing their within-class variability to that of the original dataset [23] . Table 2 presents these comparisons, demonstrating that the intra-class variances of ADASYN-generated synthetic samples closely mirrored those of their real counterparts across all classes (e.g., Class 0 Real Variance: 11340.96 vs. Synthetic Variance: 11336.63). This robust agreement in intra-class spread serves as strong evidence that ADASYN effectively maintains the inherent variability and statistical distribution of the original physiological dataset.

*Table 2. Intra-Class Variance Comparison*

| Class | Real Variance | Synthetic Variance |
|-------|---------------|--------------------|
| Class 0 | 11340.9553 | 11336.6263 |
| Class 1 | 11217.9529 | 11222.4882 |
| Class 2 | 11293.9513 | 11301.4515 |
| Class 3 | 11192.5288 | 11199.8583 |

### 2.5.2. Distribution Alignement Metrics: Fréchet Inception Distance

To quantitatively assess the fidelity of the synthetic data's distribution to that of the real dataset, a Fréchet Inception Distance (FID) was computed. This metric serves to validate the physiological plausibility of the generated samples by confirming the alignment of their underlying probability distributions. The Fréchet Inception Distance (FID) score, which measures the distance between the feature distributions of real and synthetic samples [24] was 19.8742. In image-based domains, an FID score around 20 is widely accepted as indicative of good alignment and high generative quality [25]. This reinforces the substantial similarity between the real and synthetic data distributions.

### 2.5.3. Visual Comparison via Radar Chart

The radar chart in Figure 3 illustrates the intra-class variance across the four classes for both real and ADASYN-generated synthetic ECG features. The near-perfect alignment of the two plotted polygons indicates a strong statistical resemblance, confirming that the synthetic data preserves the original distribution characteristics.

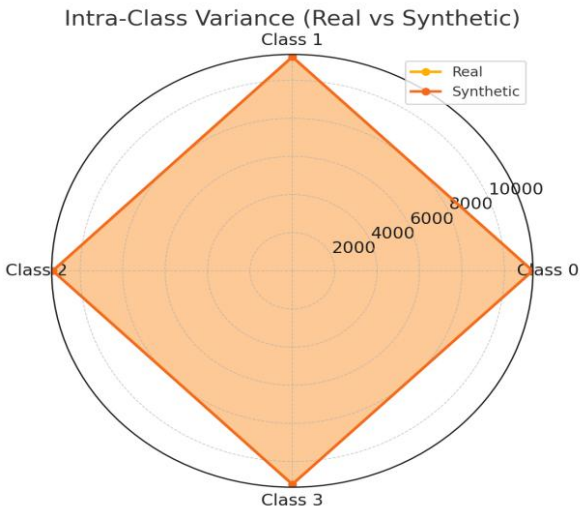

*Figure 3. Radar Chart Comparing Intra-Class Variance Between Real and ADASYN-Generated Synthetic Scalograms. The chart illustrates how closely the synthetic samples (orange line) match the real ECG scalogram feature distributions (yellow line) across all four classes. The near-perfect alignment indicates that ADASYN preserves within-class diversity, supporting the quality and fidelity of the generated samples.*

### 2.5.4. Validation Through Improved Downstream Model Performance

The ultimate practical validation of the synthetic samples' physiological plausibility and utility was demonstrated through their impact on the performance of a classification model. This was revealed on the 2DCNN unimodal model. As shown in Table 3, the observed performance gains of the model serve as compelling evidence that the synthetic samples are indeed physiologically plausible and beneficial for enhancing the model's ability to interpret and classify real-world physiological data.

*Table 3. Performance Comparison Before and After ADASYN on 2D-CNN Model*

| Metric | Before ADASYN | After ADASYN |
|--------|---------------|--------------|
| Accuracy | 82.26% | 91.94% |
| Precision | 0.84 | 0.92 |
| Recall | 0.82 | 0.92 |
| F1-Score | 0.81 | 0.92 |
| Macro Avg F1 | 0.80 | 0.91 |
| Weighted Avg F1 | 0.81 | 0.92 |

Substantial evidence, including intra-class variance comparisons, quantifiable Fréchet Inception Distance, and visual validation, demonstrated that the ADASYN-generated synthetic dataset was indistinguishable from the original. Its performance in downstream models further substantiated this assertion.

### 2.6. Denoising and Filtering

ECG signals, by their very nature, are highly susceptible to noise and artifacts [26]. These unwanted disturbances can stem from various sources, including patient movement during examination, inadequate electrode contact, or even electrical interference from surrounding equipment. If left unaddressed, this noise can severely compromise the performance of a machine learning classifier. Specifically, it often leads to overfitting [27]; a scenario where a model performs exceptionally well on the training data but fails to generalize and performs poorly on unseen testing data.

To combat this, a crucial pre-processing step was implemented before any feature extraction. The ECG dataset underwent a rigorous denoising and filtering process. The goal was to eliminate both low-frequency drift (often caused by baseline wander) and high-frequency noise/artifacts. This meticulous filtering ensures that only the most relevant physiological information remains. Specifically, frequencies within the 0.5 to 45 Hz range [28] were retained. This particular frequency band is critical because comprehensive studies have demonstrated that the diagnostically significant components of an ECG: the P-wave, QRS complex, and T-wave, all fall within this spectrum [29]. By focusing on these essential frequencies, we aimed to provide a cleaner, more reliable dataset for subsequent model training.

### 2.7. Feature Representation

Studies have shown that effective feature representation is an essential, crucial, and necessary requirement for the optimum performance of deep learning models [30]. In general, features dictate the quality of information, which is critical for accurate pattern recognition, knowledge discovery, and prediction. Domain-level feature extraction is based on the traditional signal processing techniques [31]. Features are extracted in the time, frequency, and time-frequency domains. Handcrafted statistical, morphological, and spectral characteristics are captured at this level. Because these features are domain-specific, studies have shown that they provide insights into the ECG signal dynamics [32]. At the deep learning level, the internal representations learned by neural networks were explored. This process is known as deep-feature extraction [33]. The high-level patterns and complex relationships within the data were captured from the intermediate layers of the trained models. This strategy is expected to enhance a classifier's accuracy and generalization. [34].

**Time-Domain Features:** These features are derived directly from the raw signal and represent the temporal characteristics of the data. Some features are unique to the time domain. These features include the mean signal amplitude, variance, peak-to-peak intervals, and statistical moments [35]. It has been shown that conditions, such as arrhythmias, in an ECG dataset can be identified in the time domain [36]. The time-domain feature extraction methodology comprises four sequential stages: (a) signal preprocessing, (b) feature extraction, (c) segmentation, and (d) window sliding and feature vector assembly.

The time-domain features are the first layer of information examined. Compared with the frequency or time-frequency domains, time-domain features are easier to compute. They are computationally straightforward and can reveal details of the physiological origins of the signal. The pseudocode for time-domain feature extraction is presented in Algorithm 1.

*Algorithm 1. Extraction and Preparation of 1D ECG Signals from Images*

Inputs
• D  //Directory of labeled ECG image files
• n_classes  //Number of unique class labels (e.g., 4)

Outputs

• X_train, X_val  //Preprocessed and reshaped 1D ECG training and validation sets
• y_train, y_val  //Corresponding one-hot encoded label sets

Steps

1. Initialize Storage
   X_signals ← []
   y_labels ← []

2. Extract 1D ECG Signals from Images
   For each image $f_i \in D$:
       img ← cv2.imread($f_i$, cv2.IMREAD_GRAYSCALE)
       img ← cv2.resize(img, (128, 128))
       signal ← img.mean(axis=1)       // average pixel intensity per row
       label ← get_label_from_filename($f_i$)

3. Store Processed Signals and Labels
   Append signal to X_signals
   Append label to y_labels

4. Convert to NumPy Arrays
   X_signals ← np.array(X_signals)
   y_labels ← np.array(y_labels)

5. Normalize 1D ECG Signals
   X_signals ← X_signals ÷ max(|X_signals|)       // Min-Max normalization

6. One-Hot Encode Labels
   y_encoded ← to_categorical(y_labels, num_classes = n_classes)

7. Shuffle and Split Data
   (X_train, X_val, y_train, y_val) ← train_test_split(X_signals, y_encoded, test_size = 0.2, shuffle = True)

8. Apply ADASYN Oversampling (Training Set Only)
   y_train_flat ← argmax(y_train, axis=1)
   (X_train_adasyn, y_train_adasyn) ← ADASYN().fit_resample(X_train, y_train_flat)
   y_train ← to_categorical(y_train_adasyn, num_classes=n_classes)

9. Reshape for Conv1D Input Format
   X_train ← X_train_adasyn[..., np.newaxis]
   X_val ← X_val[..., np.newaxis]

The mean, or average, is one of the most informative statistical measures derived from a signal [37]. In this experiment, the mean of the signal segments was used as the baseline measure for the overall signal level. During segmentation, the ECG signal was divided into overlapping or non-overlapping windows of equal length [38]. Relevant features were computed for each window. As shown in Algorithm 1, the extracted features were stacked and stored as Numpy arrays. This serves as a representation of the complete temporal dynamics of a signal. Normalization was subsequently performed to ensure that all the features had a uniform scale. Thus, the risk of overfitting [39] during the modeling phase was reduced.

**Frequency-Domain Features:** Periodic components within the signal were analyzed using frequency domain analysis. [40]. The frequency-domain features are particularly useful for understanding periodicity and differentiating between signals [41]. The time domain may not readily reveal the similarities between closely spaced signals; however, examination in the frequency domain often reveals a distinct pattern. Feature extraction in the frequency domain consists of four steps: a) frequency transformation, b) computation of key features, c) normalization, and d) construction of feature vectors. As listed in Table 4, a Fast Fourier Transform (FFT) was used to convert segments of the ECG signals from the time domain to the frequency domain [42]. A spectrum showing the amplitude or power of the signal at different frequencies is identified. The frequency of the signal is displayed as a function of time [43]. The FFT provides a time-frequency matrix, where each element corresponds to the power of the signal at a particular frequency and time [44]. Normalization was performed to ensure standardization.

*Table 4. Implementation steps of extracting features at the frequency domain in Python*

| Task | Description | Python/keras |
|---|---|---|
| Initialize Storage | Create empty lists `X_fft` and `y_labels` | List initialization (`[]`) |
| Extract FFT Features | Compute the FFT of 1D ECG signals (see Algorithm 1). | fft(signal, n=fft_bins), np.abs(fft(signal)) |
| Normalize FFT Features | Scale FFT features using StandardScaler. | scaler.fit_transform(fft_features.reshape(-1, 1)) |
| Store FFT Features | Save extracted FFT features in an array. | X_fft.append(fft_features) |
| Convert FFT Features to NumPy Array | Ensure FFT features are in NumPy array format. | np.array(X_fft) |
| Train-Test Split for FFT Features | Split FFT dataset into training and test sets. | train_test_split(X_fft, y_labels, test_size=0.2, stratify=y_labels) |
| One-Hot Encode Labels for FFT Features | Convert categorical labels into one-hot format. | tf.keras.utils.to_categorical(y_train, num_classes=4), tf.keras.utils.to_categorical(y_test, num_classes=4) |

Normalization was performed to account for variations in signal magnitude, thereby making the model less sensitive to amplitude differences across recordings. The frequency-domain features were transformed into a feature vector that represented each segment. Table 4 outlines the implementation steps for extracting features in the frequency domain using Google Colab (Python/Keras).

**Time-Frequency Features:** Time-frequency analysis combines the advantages of both the time and frequency domains by capturing the evolution of the signal's frequency content [45]. The spectrograms provide a dynamic two-dimensional view of the signal [46]. These features reveal transient events, nonstationary patterns, and localized frequency changes that time- or frequency-domain analysis alone cannot detect. Feature extraction in the time-frequency domain consists of four steps: a) time-frequency representation generation, b) feature extraction from time-frequency representations, c) feature vector construction, and d) normalization. Table 5 shows the implementation steps for extracting features in the time-frequency domain in Python/Keras.

*Table 5 Implementation steps of extracting features at the time-frequency domain in Python*

| Task | Description | Python/keras |
|---|---|---|
| Initialize Storage | Create empty lists `X_scalograms` and `y_labels` | List initialization (`[]`) |
| Extract 1D ECG Signal | Convert image into a 1D signal by computing the mean pixel intensity along the vertical axis (see Algorithm 1. | Feature Extraction (`img.mean(axis=1)`) |
| Generate Scalogram | Apply wavelet transform on the extracted 1D ECG signal to obtain a time-frequency representation (scalogram). | Wavelet Transform (`generate_wavelet_scalogram()`) |
| Store Features & Labels | Append the scalogram to `X_scalograms` and store the corresponding class label in `y_labels`. | Data Storage (`list.append()`) |
| Convert to NumPy Arrays | Convert the lists into NumPy arrays and reshape `X_scalograms` to a format compatible with CNN input (`batch_size × img_size × img_size × 1`). | Data Conversion & Reshaping (`np.array().reshape()`) |
| Return Processed Data | Return the processed time-frequency features (`X_scalograms`) and labels (`y_labels`) as outputs. | Function Return Statement (`return X, y`) |

## 2.7. Deep Feature Extraction

As outlined in Section 2.6, feature representations were extracted across three key domains: time, frequency, and time-frequency. This ensures a comprehensive capture of the signal characteristics. In addition to the conventional feature extraction from raw signal data, 'deep feature' extraction was employed [47]. This technique leverages the internal representations learned by training deep-learning models. Deep-learned features capture complex patterns and relationships within an ECG dataset. They are highly informative for downstream analysis and enhance the feature representation by adding details and granularity [48]. The extracted features were systematically analyzed to determine their contributions to the decision-making process. In this study, 1D-CNN, 2D-CNN, and 1D-CNN-Transformer learning algorithms were used for deep feature extraction.

### 2.8. Hypothesis and Theory Development

The next phase of the experiment focused on enhancing the classification performance of an ECG model by fusing features from multiple domains at an intermediate level. The following questions were answered.

a. Does domain fusion enhance the predictive capability of ECG deep learning model?
b. Is the complementarity of features necessary for achieving optimal performance in a multimodal ECG deep learning architecture?

These research questions were answered using five ECG classifiers: three unimodal and two multimodal.

- Hypothesis

*While complementary feature domains enhance classification performance in hybrid multimodal deep learning models, incorporating redundant or non-discriminative domains increases complexity and suboptimal performance.*

Neglecting complementarity can introduce redundant or noisy features, resulting in a) overfitting, b) higher computational demands, and c) poorer generalization of multimodal ECG deep learning architectures. Figure 4 shows a flowchart of multimodal ECG deep learning. Using the information complementarity principle, this study aimed to probe the performance of a multimodal deep learning model by adding diverse representations.

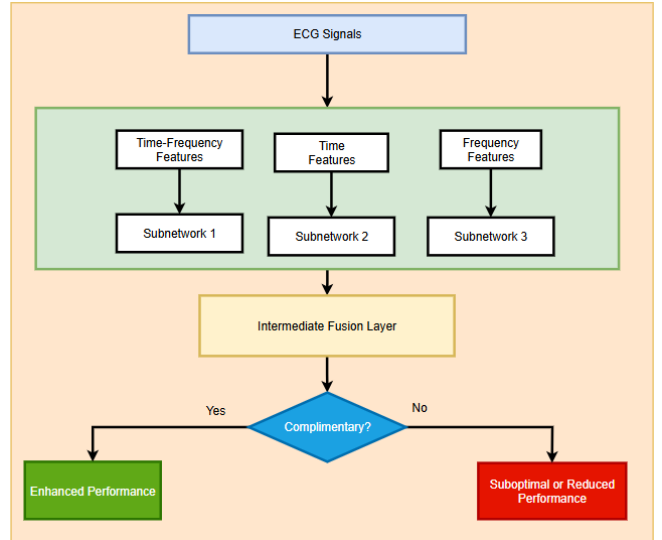

*Figure 4. Complementary Feature Domains for Optimal Multimodal Deep Learning Performance Flowchart.*

This hypothesis was investigated through empirical analysis, statistical analysis, ablation studies, and scientific reasoning.

#### A. EMPIRICAL ANALYSIS

The 1DCNN (time-domain features), 2DCNN (time-frequency domain features), and 1DCNN-Transformer (frequency-domain features) were designed, developed, and evaluated individually. Their deep features were derived to determine their complementarity.

a. 1D Convolutional Neural Network (1D-CNN) for Time Domain ECG Feature Extraction

A 1D Convolutional Neural Network (1D-CNN) was trained to analyze features extracted from the time domain. The model learned meaningful temporal patterns directly from raw waveform data. Convolutional filters were deployed to slide across the time-series data. The convolutional filter can detect P-waves, QRS complexes, and T-waves with their corresponding durations and amplitudes in an ECG signal. The 1D-CNN architecture, outlined in Table 7, consists of three convolutional blocks.

The first block utilized 64 filters, with the number doubling in each successive block. A gradual increase in the number of filters was deliberately performed to reduce overfitting. Max Pooling with a size of 2 was implemented in each layer to maximize feature extraction. It also reduces the dimensionality of the feature space. Categorical cross-entropy was used for convergence [49]. Accelerated learning and adaptive learning rates were executed using the Adam optimizer [50]. The 1D-CNN effectively captured the essential local temporal patterns from the ECG waveforms [51]. This makes the models highly effective in extracting the necessary and sufficient features for modeling. The features extracted from the 1D-CNN model are shown in Figure 5.

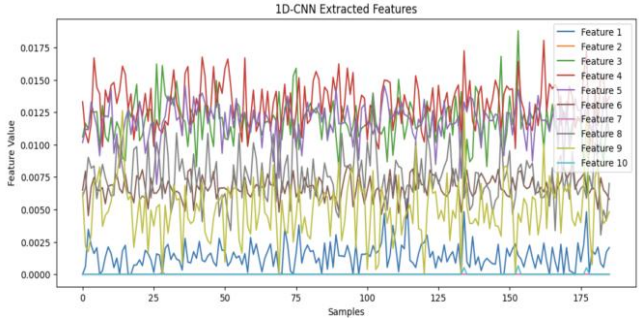

*Figure 5. 1D-CNN Extracted Features*

*Table 6 Architecture of 1D-CNN*

| Component | Detailed Description |
|---|---|
| **Input Layer** | **Input Shape:** (X_train_1d_balanced.shape[1], 1) **Purpose:** Accepts 1D ECG signals for time-domain feature extraction. |
| **1st Convolutional Block** | **Layers:** Conv1D(64, kernel_size=7, relu, padding='same', l2=0.001), BatchNormalization(), MaxPooling1D(pool_size=2) **Purpose:** Extracts local patterns while normalizing activations for stability. |
| **2nd Convolutional Block** | **Layers:** Conv1D(128, kernel_size=7, relu, padding='same', l2=0.001), BatchNormalization(), MaxPooling1D(pool_size=2) **Purpose:** Expands feature extraction capacity with increased filters. |
| **3rd Convolutional Block** | **Layers:** Conv1D(256, kernel_size=5, relu, padding='same', l2=0.001), BatchNormalization(), MaxPooling1D(pool_size=2) **Purpose:** Further enhances hierarchical feature extraction while reducing spatial dimensions. |
| **Flatten Layer** | Converts extracted features into a 1D representation for dense layers. |
| **Fully Connected Layer** | **Layers:** Dense(512, relu, l2=0.001), Dropout(0.5) **Purpose:** Introduces high-level feature abstraction while preventing overfitting. |
| **Output Layer** | **Layer:** Dense(num_classes, softmax) **Purpose:** Multi-class classification of ECG signals. |
| **Model Compilation** | **Optimizer:** Adam (learning rate = 0.0001) **Loss Function:** categorical_crossentropy **Evaluation Metric:** accuracy. |

As outlined in Figure 5, the 1D-CNN model's visual output features comprise a total of ten distinct features across approximately 190 ECG samples. Each feature represents a unique learned representation of the ECG signals after they have passed through the convolutional layers and the activation functions of the neural network. The figure shows that features 1 to 9 exhibited significant variability across the different ECG samples. This variability suggests that the features span the various classes of ECG signals. Feature 10 is an exception, with a very low magnitude. This suggests that Features 1 and 9 are the most important features of the model. Feature 10 was potentially redundant. The extracted features are critical for the accurate classification of the ECG signals.

The 1D-CNN model was trained and evaluated. The accuracy and loss graph results are shown in Figure 6.

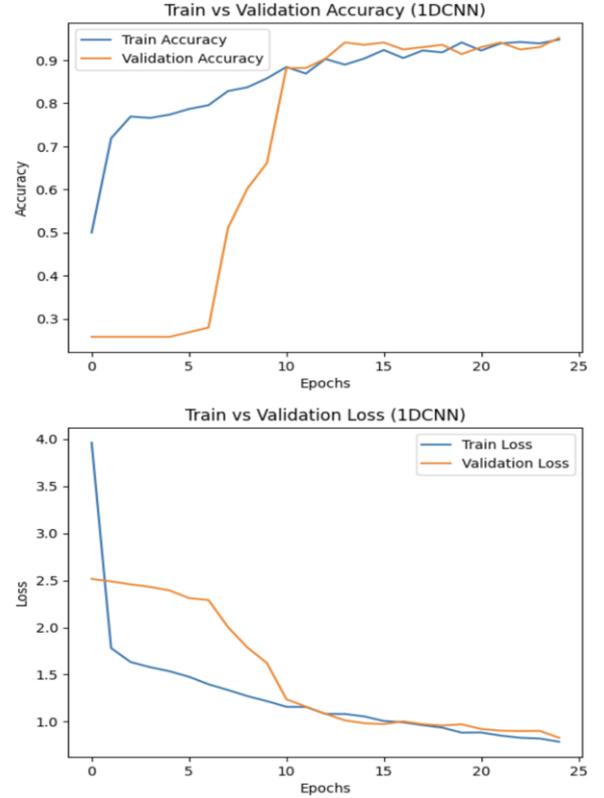

*Figure 6. 1D-CNN Accuracy and Loss Graphs As shown in*

The model has an accuracy of 0.94

b. 2D Convolutional Neural Network (2D-CNN) for Time-Frequency Domain ECG Feature Extraction

A separate 2D convolutional network was trained on the spectrograms obtained from the time-frequency domain. The uniqueness of the 2D-CNN model lies in its ability to extract features simultaneously, at both time and frequency. The architecture of the 2D-CNN model is listed in Table 7.

*Table 7. Architecture of 2D-CNN*

| Component | Detailed Description |
|---|---|
| Input Layer | **Input Shape:** (img_size, img_size, 1) **Purpose:** Accepts grayscale 2D ECG images for feature extraction. |
| 1st Convolutional Block | **Layers:** Conv2D(32, (3,3), relu, padding='same'), BatchNormalization(), MaxPooling2D(pool_size=(2,2)) **Purpose:** Captures low-level spatial features while normalizing activations for stability. |
| 2nd Convolutional Block | **Layers:** Conv2D(64, (3,3), relu, padding='same'), BatchNormalization(), MaxPooling2D(pool_size=(2,2)), Dropout(0.2) **Purpose:** Extracts deeper spatial representations and applies dropout to prevent overfitting. |
| 3rd Convolutional Block | **Layers:** Conv2D(128, (3,3), relu, padding='same'), BatchNormalization(), MaxPooling2D(pool_size=(2,2)) **Purpose:** Learns higher-order spatial hierarchies. |
| 4th Convolutional Block | **Layers:** Conv2D(256, (3,3), relu, padding='same'), BatchNormalization(), MaxPooling2D(pool_size=(2,2)) Purpose: Captures fine-grained patterns for deeper feature extraction. |
| Flatten Layer | Converts multi-dimensional feature maps into a 1D vector for dense layers. |
| Fully Connected Layer | **Layers:** Dense(512, relu), Dropout(0.5) **Purpose:** High-level feature transformation while mitigating overfitting. |
| Output Layer | **Layer:** Dense(num_classes, softmax) **Purpose:** Multi-class classification of ECG image-based representations. |

As shown in Table 7, the model comprises convolutional layers designed to extract meaningful features from the time-frequency domain. Convolutional layers apply small learnable filters (kernels) to detect local patterns in the input, such as edges, textures, or temporal structures [52]. As illustrated in the table, the model comprises four convolutional blocks.

The first block consisted of 32, 3x3 filters. In each subsequent block, the number of filters was doubled, while the filter size remained constant at 3×3. MaxPooling selected the maximum pixel value within a 2 × 2 window. Padding='same' ensures that the output retains the same spatial dimensions as the input. ReLU was used for nonlinearity. Batch normalization and dropout were used to reduce overfitting [53]. Figure 7 illustrates eight distinct feature maps generated by the convolutional layers of the 2D-CNN model.

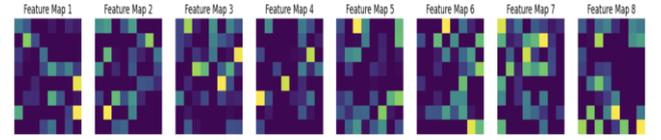

*Figure 7. 2D-CNN Extracted Feature Maps*

Each feature map reveals the spatial activation patterns that the model has learned. It captures localized spatial information, which is essential for highlighting specific regions within the input spectrogram and is crucial for discriminative classification. As highlighted in Figure 7, the maps show some density, suggesting that multiple frequency and time regions contain relevant discriminatory information. The predictive model's accuracy was 91%. Figure 8 shows the accuracy and loss graphs of the 2D-CNN model.

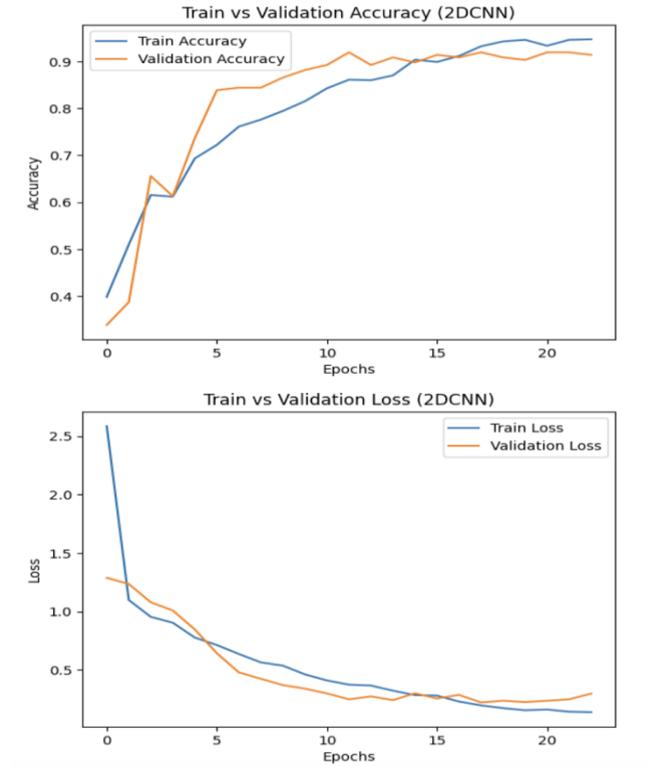

*Figure 8. 2D-CNN Accuracy and Loss Graphs*

### c. 1D Convolutional Neural Network Transformer (1D-CNN- Transformer) for Frequency Domain ECG Feature Extraction

Long-range dependency modeling of transformers was combined with the local feature extraction capabilities of 1D-CNNs [54]. In the frequency domain, the transformer can detect P-waves, QRS complexes, and T-waves in an ECG signal. The self-attention mechanism of transformers makes

8    VOLUME XX, 2017

them a perfect fit for learning the relationship between features in the ECG dataset. Its multi-head attention enables it to learn patterns at different points in a dataset [55]. The architecture of the 1D-CNN transformer model used in this study is listed in Table 8. As shown in the table, the model consists of a Transformer Block and a 1D Convolutional Neural Network (CNN) Block. Fusion of the two models (blocks) was performed using a fusion block.

*Table 8. The architecture of 1D-CNN Transformer*

**CNN-Transformer Model Description**

| Component | Detailed Description |
|---|---|
| Transformer Block (Self-Attention Mechanism) | **Input Shape**: (feature_dim, seq_length) **Purpose**: Captures long-range dependencies in frequency-domain ECG features. **Layers**: Multi-Head Attention (num_heads=12, key_dim=256) → Residual Connection (Add) → Layer Normalization → Feed-Forward Network (Dense(512, relu) → Dense(input_dim)) → Residual Connection → Layer Normalization. |
| 1D-CNN Branch (Frequency-Domain ECG features) | **Input Shape**: (X_train_fft_balanced.shape[1], 1) **Purpose**: Extracts local frequency-domain representations from FFT-transformed ECG signals. **Layers**: Conv1D(128, kernel_size=5, relu, padding='same', l2=5e-5) → MaxPooling1D(pool_size=2). |
| Feature Fusion | **Method**: Transformer block refines CNN features, capturing hierarchical relationships. **Post-processing**: Flatten → Dense(1024, relu, l2=5e-5) → Dropout(0.1). |
| Classification Layer | **Output Layer**: Dense(len(class_labels), softmax) for multi-class classification of ECG abnormalities. |
| Model Compilation Details | **Optimizer**: Adam (learning rate = 1e-4) **Loss Function**: categorical_crossentropy **Evaluation Metric**: accuracy. |

It should be noted that dropout (rate = 0.1) and L2 regularization (λ = 5e-5) were not applied inside the Transformer encoder block. Instead, it was applied only once, the Feature Fusion block. This was done to regularize the post-transformer representation before classification.

Figure 9 shows the features extracted from the 1DCNN-Transformer model.

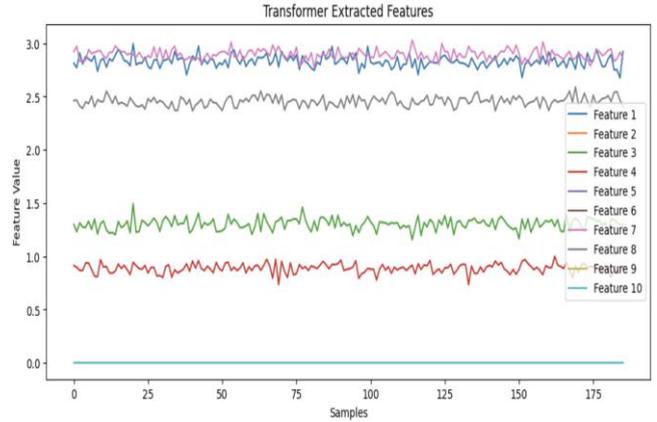

*Figure 9. 1DCNN-Transformer extracted features*

Figure 9 illustrates the transformer model's ability to leverage attention mechanisms, capture critical long-range dependencies within frequency-domain signals, and translate them into distinctive and stable representations. As shown in the diagram, the features were clearly separated into multiple distinct clusters. Features 1, 6, and 7 formed the uppermost cluster, likely representing dominant attributes. Features 3 and 4 occupied mid-level ranges (~0.8–1.5), potentially encoding secondary or supplementary temporal information relevant to classification.

The functionality of the model as a classifier, with 91% accuracy, is shown in Figure 10.

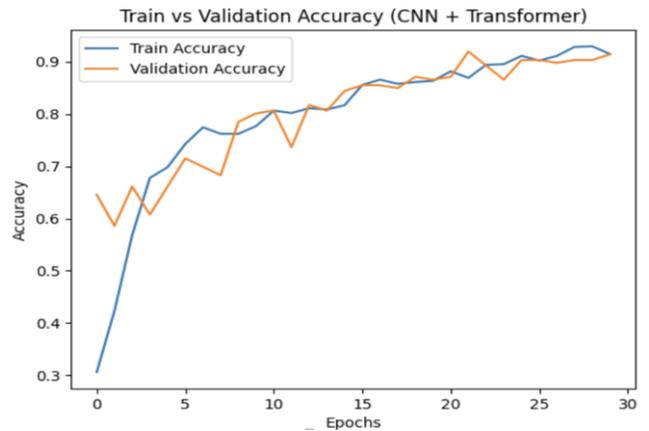

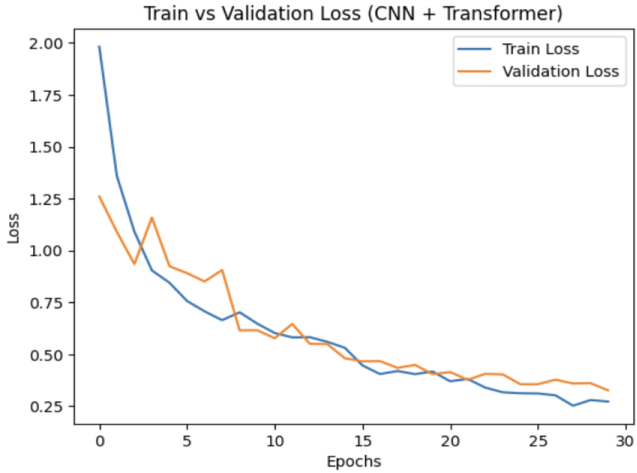

*Figure 10. Accuracy and Loss Graphs for 1DCNN-Transformer*

### d. Hybrid 1: Architectural Structure of the Hybrid 1 Multimodal Deep Learning Model (1D-CNN + 2D-CNN Features)

The 1D CNN and 2D CNN were fused, and their predictive performance was evaluated. This becomes Hybrid 1. This hybrid multimodal deep learning model combines multiple features extracted from the deep learning algorithms to classify ECG signals effectively. Table 9 provides a detailed description of the model's internal structure. As shown in the table, the model integrates two primary branches: the pretrained 1D convolutional neural network (1D-CNN) and the pre-trained 2D convolutional neural network (2D-CNN). These branches individually extract relevant features from the ECG signals represented differently, thereby enhancing the robustness and accuracy of classification.

Features extracted from the two branches through concatenation were combined, thereby effectively merging distinct temporal and spectral-temporal representations into a unified feature space. This concatenated representation passes through another dense layer. Finally, predictions were made using a dense classification layer with four neurons (four classes) and Softmax activation. This layer provides probabilistic classification results that are suitable for distinguishing between the four classes of cardiovascular diseases in the dataset. Figure 11 shows the accuracy and loss graphs for the training and validation datasets, respectively. The model's accuracy was 96%. This is an improvement over the previous unimodal models. This result suggests that fusing the features from the two models is beneficial and supports the hypothesis that complementary domains improve the performance.

*Table 9. Hybrid Multimodal Deep Learning Model Structure (Without Transformer)*

| Component | Detailed Description |
| --- | --- |
| 1D-CNN Branch (Time-Domain ECG features) | **Input Shape:** (X_train_1d.shape[1], 1) <br> **Purpose:** Extract temporal features from ECG signals using 1D convolutions. <br> **Layers:** Pre-trained 1D-CNN initial layer (model_1d.layers[0]) followed by intermediate convolutional layers (model_1d.layers[1:-2]). <br> **Post-processing:** Flatten → Dropout(0.5) → Dense(256, relu). |
| 2D-CNN Branch (Time-Frequency ECG features) | **Input Shape:** (X_train_2d.shape[1], X_train_2d.shape[2], 1) <br> **Purpose:** Capture patterns from ECG time-frequency representations (e.g., spectrogram/wavelet). <br> **Layers:** Pre-trained 2D-CNN initial layer (model_2d.layers[0]) followed by intermediate convolutional layers (model_2d.layers[1:-3]). <br> **Post-processing:** Flatten → Dropout(0.5) → Dense(256, relu). |
| Feature Fusion | **Method:** Concatenate deep features from 1D-CNN and 2D-CNN branches. <br> **Combined Representation:** Dense(256, relu) with L1 regularizer (l1=0.01). <br> **Regularization:** Dropout(0.5). |
| Classification Layer | **Output Layer:** Dense(num_classes, softmax) for multi-class ECG signal classification into cardiovascular diseases. |
| Model Compilation Details | **Optimizer:** Adam (lr=0.0001) <br> **Loss Function:** categorical_crossentropy <br> **Evaluation Metric:** accuracy. |

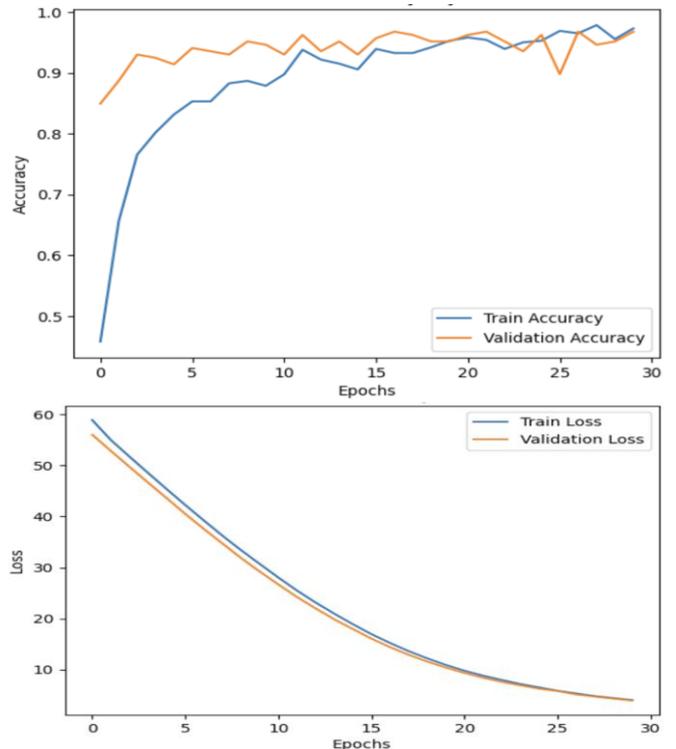

*Figure 11. Accuracy and Loss Graphs of Hybrid 1 Multimodal Deep Learning Model Structure (1D-CNN + 2D-CNN Features)*

### e. Hybrid 2: Architectural Structure of Hybrid2 Multimodal Deep Learning Model (1DCNN + 2DCNN + Transformer)

The Hybrid 2 model integrates features from the 1DCNN + 2DCNN + Transformer models, resulting in three parallel branches. Each branch processed a unique representation of the ECG signal to extract complementary features. The architectural design of the model is presented in Table 10. This arrangement allowed the model to learn more robust and comprehensive patterns.

After extracting the distinct features from the three branches, the model combines the temporal, spatial, and spatial-temporal time-frequency domains. Dropout and L1 regularizer were used to reduce overfitting. Finally, the output layer employs a dense layer consisting of four neurons, which is equal to the number of cardiovascular disease classes activated by the Softmax function. This layer outputs the classification probabilities across the four classes.

Figure 12 shows the accuracy and loss graphs for the training and validation datasets. The model's accuracy was 94%. This represents a 2% decrease in the Hybrid 1 model's result.

*Table 10. Hybrid Multimodal Deep Learning Model Structure (1D-CNN + 2D-CNN + Transformer)*

| Component | Detailed Description |
| --- | --- |
| 1D-CNN Branch (Time-Domain ECG features) | **Input Shape**: (X_train_1d.shape[1], 1) <br> **Purpose:** Extract temporal features directly from ECG signals using 1D convolutions. <br> **Layers:** Pre-trained 1D-CNN initial layer (model_1d.layers[0]) followed by intermediate convolutional layers (model_1d.layers[1:-2]). <br> **Post-processing:** Flatten → Dropout(0.5) → Dense(256, relu). |
| 2D-CNN Branch (Wavelet/Spectrogram features) | **Input Shape**: (X_train_2d.shape[1], X_train_2d.shape[2], 1) <br> **Purpose:** Capture spatial-frequency patterns from ECG spectrogram or wavelet representations. <br> **Layers:** Pre-trained 2D-CNN initial layer (model_2d.layers[0]) followed by intermediate convolutional layers (model_2d.layers[1:-3]). <br> **Post-processing:** Flatten → Dropout(0.5) → Dense(256, relu). |
| Transformer Branch (Frequency-Domain features) | **Input Shape:** (X_train_transformer.shape[1], 1) <br> **Purpose:** Extract frequency-domain patterns and long-range dependencies from ECG signals. <br> **Layers:** Custom Transformer-based model (transformer_model). <br> **Post-processing:** Transformer output embeddings directly used. |
| Feature Fusion | **Method:** Concatenate deep features from 1D-CNN, 2D-CNN, and Transformer branches. <br> **Combined Representation:** Dense(256, relu) with L1 regularizer (l1=0.01). <br> **Regularization:** Dropout(0.5). |
| Classification Layer | **Output Layer:** Dense(num_classes, softmax) for multi-class ECG signal classification into cardiovascular diseases. |
| Model Compilation Details | **Optimizer:** Adam (lr=0.0001) <br> **Loss Function:** categorical_crossentropy <br> **Evaluation Metric:** accuracy. |

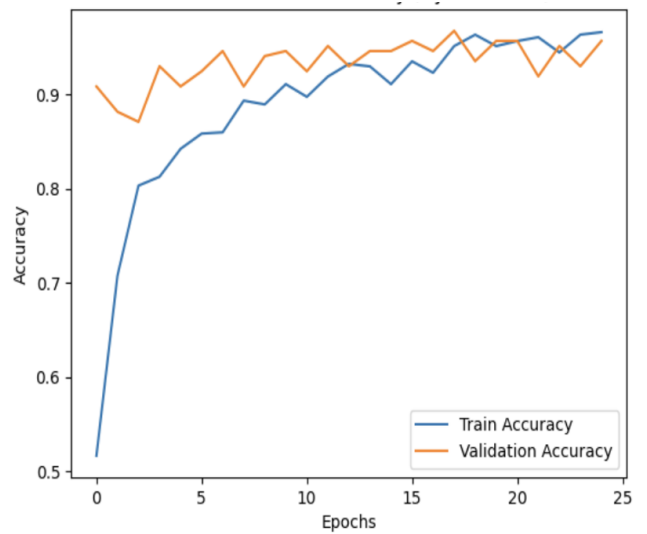

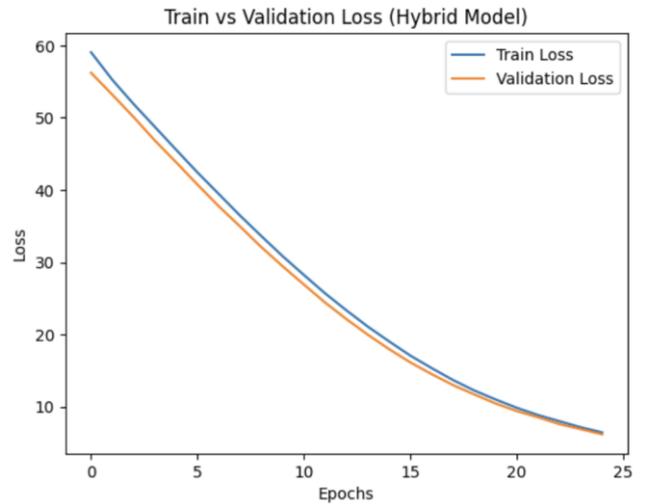

*Figure 12. Accuracy and Loss Graphs of Hybrid 1 Multimodal Deep Learning Model Structure (1D-CNN + 2D-CNN + Transformer Features)*

### f. Empirical Result

Figure 13 presents the classification accuracy, precision, recall, and F1 score obtained from the five different model configurations (three unimodal and two multimodal architectures). Among the unimodal models, the 1D-CNN trained on time-domain features achieved the highest performance in all metrics, followed by the 2D-CNN model trained on time-frequency features, and finally, the 1D-CNN transformer trained on frequency-domain features.

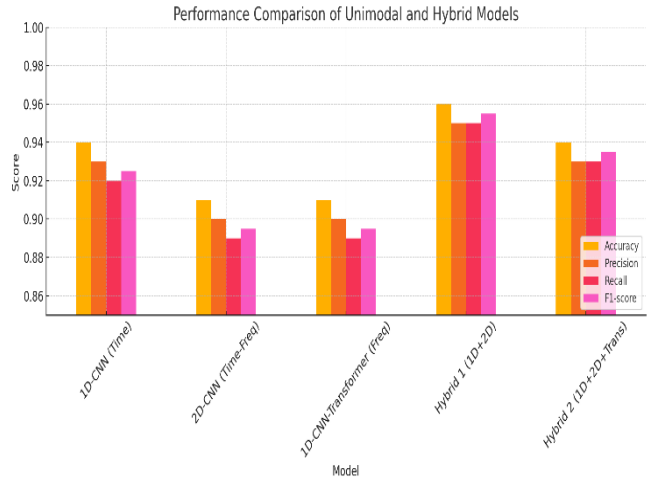

*Figure 13. Performance Comparison of Unimodal and Multimodal Models. The bar chart presents the comparative performance of the five ECG classification models across four evaluation metrics: Accuracy, Precision, Recall, and F1-score. The unimodal models: 1D-CNN (Time), 2D-CNN (Time-Frequency), and Transformer (Frequency) demonstrate varied performance, with the 1D-CNN model outperforming the other unimodal variants. Notably, the Hybrid 1 model (1D-CNN + 2D-CNN), which fuses time-domain and time-frequency features, achieves the highest performance across all metrics, with peak accuracy at 96%. In contrast, Hybrid 2 (1D-CNN + 2D-CNN + Transformer), which includes an additional frequency-domain branch, exhibits a decline in performance, indicating potential feature redundancy. These results empirically support the hypothesis that the effectiveness of fusion is driven not by the number of domains but by the **complementarity** of the fused features.*

Among the multimodal methods, Hybrid 1, which integrates features from both the 1D-CNN (time domain) and 2D-CNN (time-frequency domain), achieved the best overall performance in all metrics considered, demonstrating the strength of fusing complementary feature domains. However, when the frequency-domain deep features from the 1D-CNN transformer were added to create Hybrid 2, the performance decreased despite the increase in architectural complexity. This result suggests that the introduction of additional features introduces redundancy rather than complementarity, thereby diminishing the effectiveness of fusion.

### g. Evaluation of the Best-Performing Model (Hybrid 1)

As shown in Figure 13, the Hybrid1 model (1D-CNN + 2D-CNN) achieved the highest overall classification accuracy and F1-score among all tested architectures and fusion strategies. Table 11 presents the confusion matrix from the model's test data predictions, highlighting the number of correct and incorrect predictions for each diagnostic class. This matrix enables visual inspection of inter-class confusion, especially between clinically adjacent conditions such as 'MI' and 'HistoryMI'. Table 12 summarizes the performance metrics for each class, including Precision, Recall, F1-Score, and Specificity. These metrics provide a granular view of the model's ability to correctly identify true positives and reject false positives for each condition. Notably, the model achieved perfect recall and a very high F1-score for the 'MI' and 'Normal' classes, while maintaining strong performance for more challenging categories such as 'History of MI' and 'Abnormal heartbeat'.

*Table 11. Confusion Matrix of Hybrid 1*

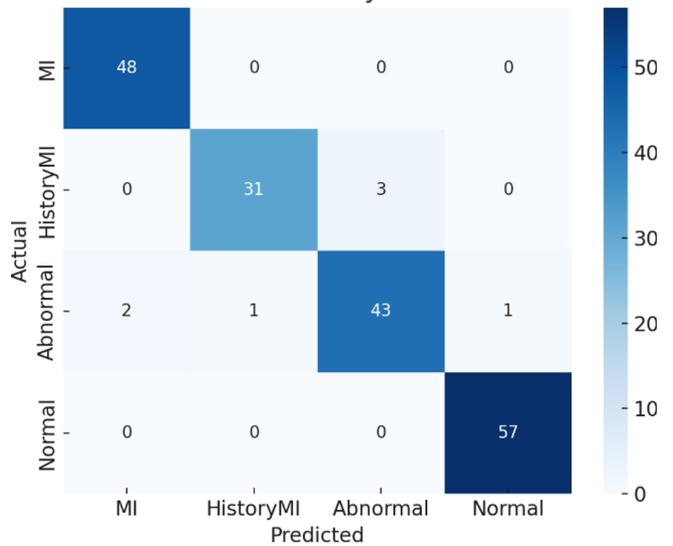

Table 12. Per-Class Metrics of Hybrid 1

| Class | Precision | Recall | F1-Score | Specificity | Support |
|---|---|---|---|---|---|
| MI | 0.96 | 1.00 | 0.98 | 0.99 | 48 |
| HistoryMI | 0.97 | 0.91 | 0.94 | 0.97 | 34 |
| Abnormal | 0.93 | 0.91 | 0.92 | 0.97 | 47 |
| Normal | 0.98 | 1.00 | 0.99 | 1.000 | 57 |
| Accuracy | | | **0.96** | | 186 |
| Macro Avg | 0.96 | 0.96 | 0.96 | 0.98 | 186 |
| Weighted Avg | 0.96 | 0.96 | 0.96 | 0.98 | 186 |

### B. STATISTICAL ANALYSIS

The performance improvement when the 1D-CNN was fused with the 2D-CNN was investigated. Additionally, the decrease in accuracy from 96% (Hybrid Model 1) to 94% (Hybrid Model 2) was examined. Correlation, mutual information, bootstrapping, and Bayesian inference were used for the statistical analysis.

#### a. Correlation and Mutual Information Analysis

Correlation [56] and mutual information analyses [57] have been used in various fields of science to analyze the statistical relationships between two or more entities. In this study, the correlation and mutual information between the features of 1D-CNN, 2D-CNN, and Transformer models were visualized using heatmaps to evaluate their relationships. This analysis provides insights into how well these features complement or interfere with one another.

Figure 14 shows a heatmap representing the correlation analysis between the feature representations from ID-CNN, 2D-CNN, and transformer models. The correlation heatmap shows a linear relationship between the features. The mutual information heatmap depicted in Figure 15 was used to show statistical dependence beyond the linearity assumption of the correlation analysis.

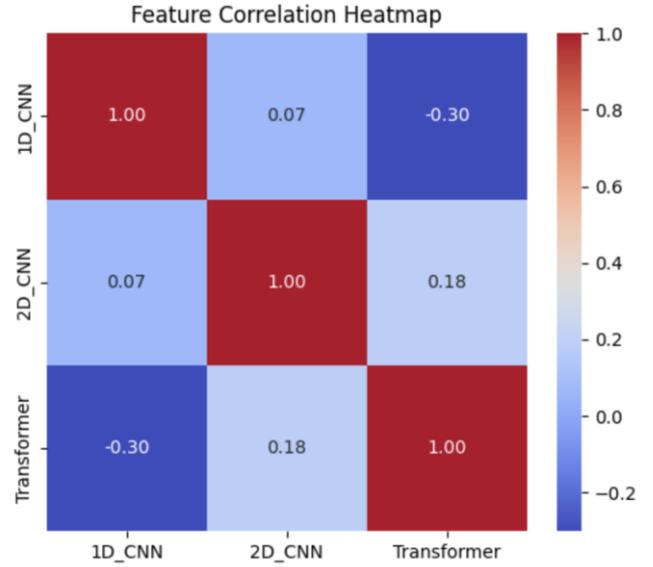

Figure 14. Feature Correlation Heatmap

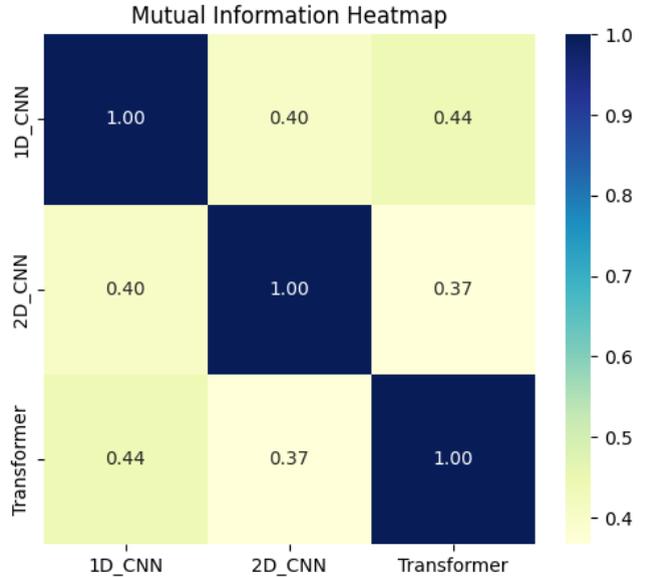

Figure 15. Feature Mutual Information (MI) Heatmap

**Comparison of Correlation and Mutual Information (MI) Heatmaps:**

- **1D-CNN vs. 2D-CNN**
    - The correlation heatmap shows a low positive correlation (+0.07), indicating minimal linear dependence.
    - However, the MI heatmap revealed a moderate MI value (0.40), suggesting the presence of nonlinear statistical dependence.
- **1D-CNN vs. Transformer:**

- The correlation heatmap showed a negative correlation (-0.30), indicating an inverse linear relationship.
- The MI heatmap showed a moderate MI value (0.44), indicating statistical dependence beyond linear patterns, regardless of negative linear correlation.

- **2D-CNN vs. Transformer:**
  - The correlation heatmap shows a weak positive correlation (+0.18), indicating a small degree of linear dependence.
  - The MI heatmap showed a moderate MI value (0.37), again indicating statistical dependence beyond linear patterns.

**Interpretation:** Using both correlation and mutual information heatmaps, this study showed that the relationships between the features were beyond linearity. The MI heatmap reveals that even when the linear correlation is weak or negative, there is still a significant statistical dependence owing to nonlinear relationships.

**Implications for Performance:**

1. **Complementarity and Non-Linearity**
   - There was a low linear correlation between the 1D-CNN and 2D-CNN, but a moderate MI value (0.40) between them. This suggests that these features contain complementary information beyond the linear patterns.
   - This nonlinear complementarity likely contributed to the improved performance of the 1D-CNN + 2D-CNN (hybrid 1) model.
2. **Conflicting Linear Patterns and Overall Dependence**
   - The correlation heat map shows a negative linear correlation (-0.30) between the 1D-CNN and transformer, indicating conflicting linear patterns, which likely hindered the performance.
   - However, the mutual information heat map showed a moderate MI (0.44). This suggests that there was still some statistical dependence. MI shows that there is still shared information between them, regardless of the inverse linear trend.
   - Therefore, the MI value is likely responsible for the influence of the transformer on model performance, even if the linear correlation is negative.
3. **Redundancy and Non-Linearity**
   - The moderate MI value (0.37) between the 2D-CNN and transformer, despite the weak linear correlation, suggests that the transformer introduces redundant information beyond the linear patterns.
   - This redundancy, both linear and nonlinear, is likely to explain why the transformer did not significantly enhance performance.

The combination of correlation and mutual information analysis provides a more comprehensive understanding of the relationships between the feature domains. While correlation captures linear dependencies, MI captures broader statistical dependencies, including non-linear relationships. Therefore, the observed performance changes in the hybrid models can be explained by considering both the linear and nonlinear dependencies among the feature domains.

b. **Bootstrapping-Based Statistical Validation**

Statistical relationships among the feature domains were further investigated using bootstrapping techniques [58]. Complementarity and redundancy analyses were also performed.

1. **Complementarity Experiment: Hybrid 1 vs 2D-CNN.**

Hybrid 1, which fuses the 1D-CNN (time-domain) and 2D-CNN (time-frequency) features, was evaluated against the 2D-CNN baseline. The goal was to determine whether combining complementary modalities enhances classification performance. Nonparametric bootstrapping (1,000 iterations with replacement) was used in the experiment. The distribution of differences across the four performance metrics (accuracy, precision, recall, and F1-score) was observed. As shown by the green dots across all the subplots in Figure 16, the bootstrapped distributions are consistently skewed above zero. This indicated that the improvements were statistically reliable. Hybrid 1 vs 2D-CNN experimentation validated the hypothesis of feature complementarity.

2. **Redundancy Experiment: Hybrid 2 vs Hybrid 1.**

The goal of this experiment was to determine the effects of the redundant domain features. Hybrid 2 vs. Hybrid 1 was examined. Transformer-based frequency-domain features were added to Hybrid 1 to obtain Hybrid 2. As shown in Figure 16, the bootstrapped metric differences (in red) for all metrics demonstrate that the distributions are tightly clustered around zero, or even slightly negative. This indicates that there is no meaningful improvement in the performance when transformer features are added. Instead, the third modality (transformer-based frequency domain) must have introduced redundant or overlapping information, which diluted the effect of the complementary signals from 1D and 2D-CNNs (Hybrid 1).

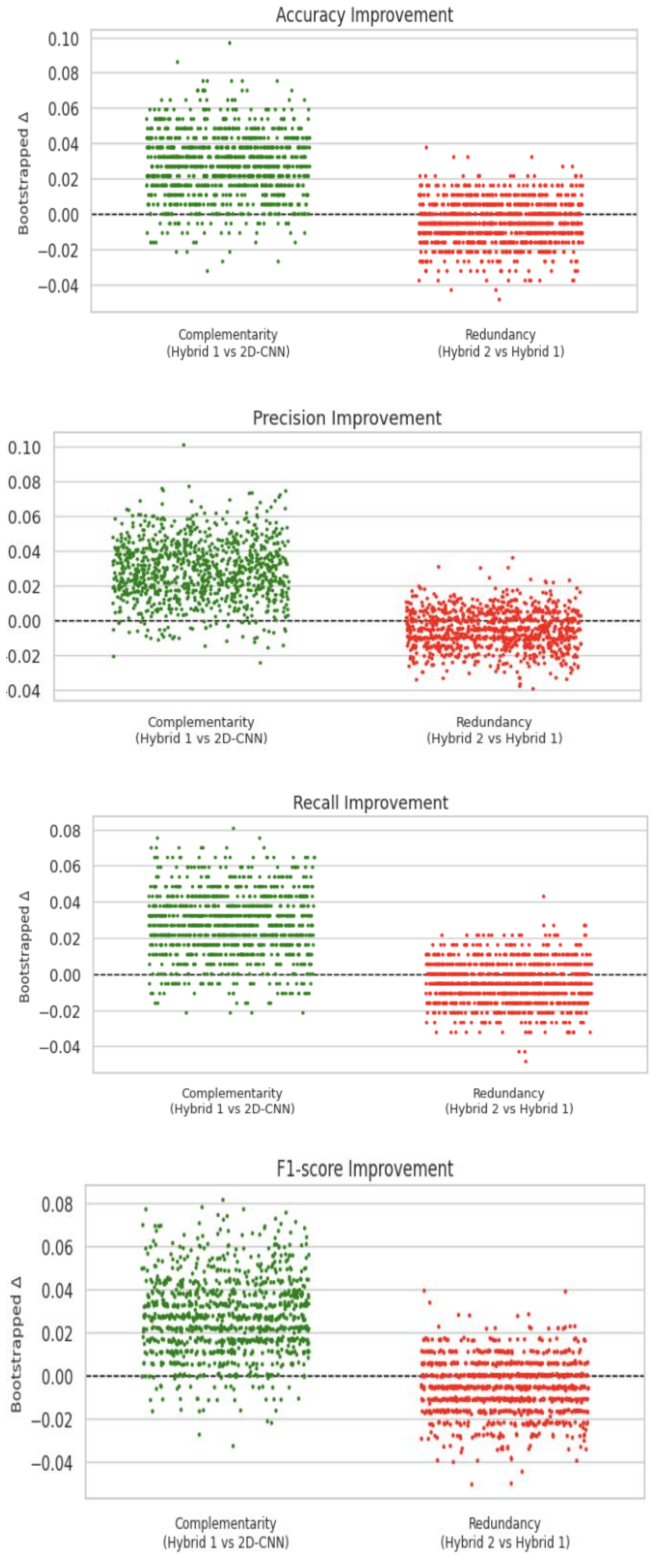

*Figure 16. Bootstrapped Metric Differences: Complementarity vs Redundancy*

To further understand the statistical significance of the improvement in the gain of the 2D-CNN when the 1D-CNN was added to Hybrid 1, a complementarity table was created, as shown in Table 13. As shown in Table 13, for all metrics, the p-values were well below the threshold of 0.05, indicating that the observed differences in performance between Hybrid 1 and 2D-CNN were statistically significant and not due to random chance. This confirms that Hybrid 1 achieves significantly better performance than the 2D-CNN as a stand-alone. Furthermore, the confidence interval did not include zero, and the mean differences were positive, suggesting a strong improvement. Therefore, the additional 1D-CNN modality provides complementary information that enhances the classification. This statistically supports the complementarity hypothesis, showing that combining time-domain and time-frequency-domain features leads to synergistic improvements in the ECG signal analysis.

*Table 13. Bootstrapping Comparison – Hybrid 1 vs 2D-CNN (Complementarity)*

| Metric | Mean Diff | 95% CI | p-value |
| --- | --- | --- | --- |
| Accuracy | 0.0497 | [0.0108, 0.0914] | 0.012 |
| Precision | 0.0413 | [0.0082, 0.0753] | 0.006 |
| Recall | 0.0477 | [0.0054, 0.0860] | 0.011 |
| F1-score | 0.0513 | [0.0101, 0.0919] | 0.008 |

Table 14 illustrates the redundancy effect of incorporating transformer frequency-domain features into Hybrid 1, resulting in Hybrid 2. As shown in the table, all the mean differences in the performance metrics are negative or close to zero, suggesting no performance gain from adding the transformer. Additionally, the 95% confidence interval was almost zero, indicating no statistically significant improvement. Furthermore, all p-values were greater than 0.05, indicating that the null hypothesis (no difference) could not be rejected.

*Table 14. Bootstrapping Comparison – Hybrid 2 vs Hybrid 1 (Redundancy)*

| Metric | Mean Diff | 95% CI | p-value |
| --- | --- | --- | --- |
| Accuracy | -0.0059 | [-0.0323, 0.0161] | 0.237 |
| Precision | -0.0047 | [-0.0272, 0.0171] | 0.328 |
| Recall | -0.0062 | [-0.0269, 0.0161] | 0.227 |
| F1-score | -0.0058 | [-0.0302, 0.0168] | 0.287 |

Considering all the indicators in the table, it is evident that the addition of the transformer model to Hybrid 1 had a negative effect. The decrease in performance after the addition of the transformer supports the redundancy hypothesis. Adding more modalities does not guarantee better performance unless the new features are complementary to the existing ones. The implementation pseudocode for bootstrapping in this study is shown in Algorithm 2.

*Algorithm 2. Bootstrap-Based Complementarity and Redundancy Analysis*

Inputs
• $\mathcal{M}_2^D, \mathcal{M}^{H1}, \mathcal{M}^{H2}$ // Trained models (2D-CNN, Hybrid 1, Hybrid 2)
• $X_2^D, X^{H1}, X^{H2}$ //Aligned test inputs for each model
• y_true //Ground truth labels: $[y_1, y_2, ..., y_n]$
• B //Number of bootstrap iterations
• $\mathcal{F} = \{f_acc, f_prec, f_rec, f\_F1\}$ //Set of evaluation metrics

Outputs
• $\mathcal{R}$ //A table of bootstrapped metric differences with labels
• Strip plot of $\Delta f$ grouped by metric and label

Steps
1. Compute model predictions
   $\hat{y}_2^D \leftarrow \text{argmax}(\mathcal{M}_2^D(X_2^D))$
   $\hat{y}^{H1} \leftarrow \text{argmax}(\mathcal{M}^{H1}(X^{H1}))$
   $\hat{y}^{H2} \leftarrow \text{argmax}(\mathcal{M}^{H2}(X^{H2}))$

2. Initialize result list
   $\mathcal{R} \leftarrow \emptyset$

3. For each metric $f \in \mathcal{F}$ do:
   // Complementarity: Hybrid 1 vs 2D-CNN
   For b = 1 to B do:
       Sample indices I_b with replacement from {1, ..., n}
       $\Delta f^b \leftarrow f(\mathbf{y}\_true[I\_b], \hat{y}^{H1}[I\_b]) - f(\mathbf{y}\_true[I\_b], \hat{y}_2^D[I\_b])$
       Append $(f, \Delta f^b, \text{"complementarity"})$ to $\mathcal{R}$

   // Redundancy: Hybrid 2 vs Hybrid 1
   For b = 1 to B do:
       Sample indices I_b with replacement from {1, ..., n}
       $\Delta f^b \leftarrow f(\mathbf{y}\_true[I\_b], \hat{y}^{H2}[I\_b]) - f(\mathbf{y}\_true[I\_b], \hat{y}^{H1}[I\_b])$
       Append $(f, \Delta f^b, \text{"redundancy"})$ to $\mathcal{R}$

4. Visualize results

### c. Bayesian-Based Statistical Validation.

As shown in section b, bootstrapping statistical analysis strongly supported the correlation analysis. Finally, a statistical investigation of the redundancy and complementarity of the feature domains on model performance was conducted using the Bayesian inference rule [59]. Bootstrapping was employed for non-parametric sampling with 1,000 iterations. Combining Bayesian inference and non-parametric bootstrapping provides a robust and interpretable analysis of model performance differences. Bayesian analysis was used to estimate the posterior distribution of metric differences across the models. Bootstrapping-derived distributions of metric differences and confidence intervals.

i. **Complementarity Experiment (Bayesian Inference): Hybrid 1 vs 2D-CNN**

The performance gain of Hybrid 1 (1D-CNN + 2D-CNN) over the 2D-CNN alone was investigated to determine whether it was statistically supported by Bayesian inference. This provided probabilistic confidence in the feature complementarity hypothesis. The accuracy, Precision, Recall, and F1-score evaluation metrics of Hybrid 1 and the 2D-CNN were assessed. The violin plot is shown in Figure 17. All metrics showed a positive mean performance gain ($\Delta > 0$), credible intervals close to or above zero, and posterior probabilities $> 0.90$. This strongly supports the fact that Hybrid 1 benefited from complementary information between the 1D and 2D CNN modalities. The performance improvement of Hybrid 1 over 2D-CNN was not negligible.

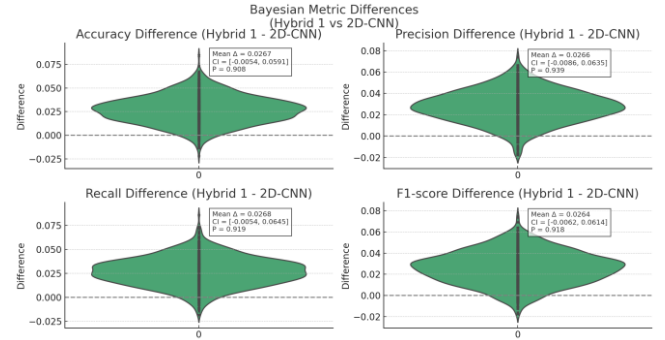

*Figure 17. Bayesian Metric Difference (Hybrid 1 vs 2D-CNN*

Table 15 presents the Bayesian comparison performances of Hybrid 1 and the 2D-CNN. The table shows a positive mean difference in all metrics considered and a confidence interval close to zero, with a posterior probability greater than 90%. Again, this confirms the complementarity of the 1D-CNN with the 2D-CNN.

*Table 15. Bayesian Comparison – Hybrid 1 vs 2D-CNN (Complementarity)*

| Metric | Mean Diff | 95% CI | P(H1 > 2D-CNN) |
|---|---|---|---|
| **Accuracy** | 0.02675 | [-0.00538, 0.05914] | 0.908 |
| **Precision** | 0.02662 | [-0.00863, 0.06352] | 0.939 |
| **Recall** | 0.02684 | [-0.00538, 0.06452] | 0.919 |
| F1-score | 0.02639 | [-0.00622, 0.06143] | 0.918 |

### ii. Redundancy Experiment (Hybrid 2 vs Hybrid 1)

The effect of redundancy was further investigated by analyzing the effect of the additional transformer (frequency domain) component in Hybrid 2. The experimental design is presented in Table 16. Models A and B were compared to determine whether they outperformed each other or not. The performance metrics used included the accuracy, precision, recall, and F1 scores. The violin plot of the results is shown in Figure 17.

*Table 16. Redundancy Experiment Setup (Hybrid 2 vs Hybrid 1)*

| Element | Details |
|---|---|
| Model A (Baseline) | Hybrid 1 (1D-CNN + 2D-CNN) |
| Model B (Test) | Hybrid 2 (1D-CNN + 2D-CNN + Transformer) |
| Comparison Goal | Check if Hybrid 2 outperforms Hybrid 1 ($\Delta$ = Metric_H2 – Metric_H1) |
| Metrics | Accuracy, Precision, Recall, F1-score |
| Sample Size | All test samples from ECG dataset |
| Evaluation Strategy | Bayesian estimation using bootstrapped samples (1,000 iterations) |
| Credible Interval | 95% Bayesian CI for each metric difference |
| Posterior Probability | $P(\Delta > 0)$, i.e., probability that Hybrid 2 is better than Hybrid 1 |

Figure 18 shows that in all the metrics considered, Hybrid 2 underperformed Hybrid1. In addition, the posterior probabilities, $P(\Delta>0)$, for each metric were well below 0.5, (0.227 - 0.328). This suggests that the transformer must have introduced redundancy that degrades or stagnates performance. The experimental outcome further shows that multimodal deep learning architectures benefit not from more domains, but from domains that offer non-overlapping complementary features.

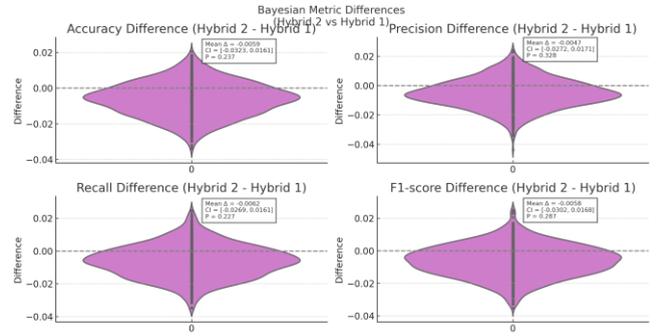

*Figure 18. Bayesian Metric Difference (Hybrid 1 vs Hybrid 2)*

A comparison of the results is presented in Table 17. The table shows that the addition of the transformer model is not beneficial. For all metrics considered (accuracy, precision, recall, and F1 score), the mean differences were negative and the probabilities of Hybrid 2 being greater than those of Hybrid 1 are well below 0.5.

*Table 17. Bayesian Comparison – Hybrid 2 vs Hybrid 1 (Redundancy)*

| Metric | Mean Diff | 95% CI | P(H2 > H1) |
|---|---|---|---|
| Accuracy | -0.00595 | [-0.03226, 0.01613] | 0.237 |
| Precision | -0.00470 | [-0.02718, 0.01712] | 0.328 |
| Recall | -0.00617 | [-0.02688, 0.01613] | 0.227 |
| F1-score | -0.00582 | [-0.03024, 0.01683] | 0.287 |

Algorithm 3 presents the steps of the Bayesian algorithm designed for this study, which was implemented in Python using the Keras library.

*Algorithm 3. Bootstrap-Based Bayesian Comparison of Model Performance*

Inputs
• $\mathcal{M}_2{}^D$, $\mathcal{M}^{H1}$ // Trained models: 2D-CNN and Hybrid 1
• $X_2{}^D$, $X^{H1}$ // Test inputs aligned to respective models
• y_true // Ground truth class labels: $[y_1, y_2, ..., y_n]$
• B // Number of bootstrap iterations
• $\mathcal{F}$ = {$f_a$cc, $f_p$rec, $f_r$ec, $f$_F1} // Evaluation metrics (weighted average)

Outputs
• $\mathcal{R}$ // Summary table with mean $\Delta f$, 95% CI, and P($\Delta f > 0$)
• Violin plots of metric difference distributions

**Steps**

1. Load Models
   $\mathcal{M}_{2D}$ ← load_model("2d_cnn_model.h5")
   $\mathcal{M}_{H1}$ ← load_model("hybrid_model_1.h5")

2. Load Test Data
   $\mathbf{X}_{2D}$ ← np.load("X_test_2d.npy")
   $\mathbf{X}_{H1}$ ← np.load("X_test_1d.npy")
   y_true ← np.load("y_test.npy")

3. Predict Model Outputs
   $\hat{y}_{2D}$ ← argmax($\mathcal{M}_{2D}(\mathbf{X}_{2D})$)
   $\hat{y}_{H1}$ ← argmax($\mathcal{M}_{H1}([\mathbf{X}_{H1}, \mathbf{X}_{2D}])$)

4. Define Bootstrap Comparison Function
   Define bootstrap_samples(fn, y, y_pred_A, y_pred_B, B):
      For each iteration $b = 1$ to **B**:
         Sample indices **I_b** ~ Uniform(1, ..., n) with replacement
         $\Delta f^b$ ← fn(y[**I_b**], y_pred_A[**I_b**]) − fn(y[**I_b**], y_pred_B[**I_b**])
         Append $\Delta f^b$ to list

5. Set Evaluation Metrics
   $\mathcal{F}$ = {
      $f_{acc}$: weighted accuracy,
      $f_{prec}$: weighted precision,
      $f_{rec}$: weighted recall,
      $f\_F1$: weighted F1-score
   }

6. Run Bayesian Comparison
   For each metric $f \in \mathcal{F}$:
      $\Delta f$ ← bootstrap_samples($f$, **y**_true, $\hat{y}^{H1}$, $\hat{y}_{2D}$, B)
      Compute summary statistics:
         μ ← mean($\Delta f$)
         $CI_{95}$ ← [2.5th percentile, 97.5th percentile] of $\Delta f$
         $P$ ← proportion($\Delta f > 0$)
         Append (metric name, μ, $CI_{95}$, $P$) to $\mathcal{R}$

7. Create Results Table
   $\mathcal{R}$ ← pd.DataFrame(columns = ["Metric", "Mean $\Delta f$", "95% CI", "P($\Delta f > 0$)"])

8. Visualize Metric Distributions
   For each $f \in \mathcal{F}$:
      Plot sns.violinplot(data = $\Delta f$, label = metric name)

### C. ABLATION STUDIES

Statistical analysis validated the empirical result that complementary features improve model performance, whereas redundant features plateau or decrease it. This concept was further investigated in an ablation study. The ablation study presented in Figure 19 evaluated complementarity versus redundancy.

The study was based on a comparison of all models in the empirical analysis: 1D-CNN (time domain, 94%), 1D-CNN-Transformer (frequency domain, 91%), and 2D-CNN (time-frequency domain, 91%), with two hybrid approaches: Hybrid 1 (combining 1D-CNN and 2D-CNN) and Hybrid 2 (combining 1D-CNN, 2D-CNN, and transformer). The results demonstrated that Hybrid 1 achieved the highest accuracy of 96%, which is an improvement of 2% over the best unimodal baseline (1D-CNN). This significant gain suggests a strong complementarity between the time and time-frequency domain features. Conversely, Hybrid 2, which incorporated additional transformer-based frequency-domain features, showed no gain over the baseline accuracy, achieving the same accuracy of 94% as that of the best unimodal model. This indicates potential redundancy or suboptimal integration of the frequency-domain features provided by the transformer model.

Overall, this ablation illustrates the importance of carefully balancing the feature complementarity and redundancy to optimize the performance of the hybrid model in ECG classification tasks.

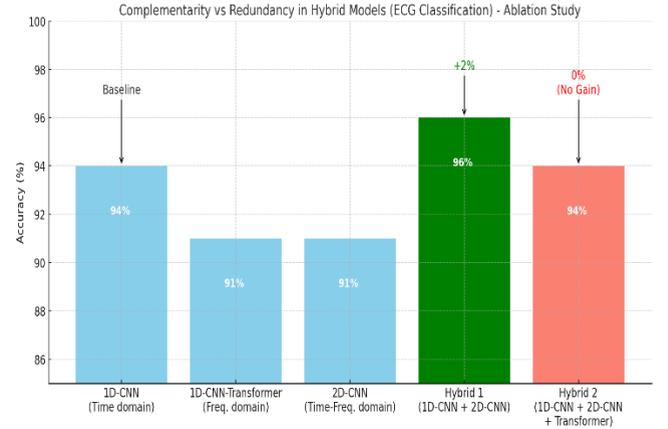

*Figure 19. Ablation Study Chart*

### D. SCIENTIFIC REASONING FRAMEWORK: MULTIMODAL FEATURE DEPENDENCE AND PERFORMANCE SYNERGY

Neither empirical nor statistical analyses nor ablation studies have provided evidence against this hypothesis. Therefore, the phenomenon was established using a scientific reasoning framework.

a) Theoretical Premises (Mathematical Statements)
- Premise 1 (Linear Independence and Complementarity)
  - Let $D_i$, $D_j$ be the feature domains ($D_i$ denotes the *i-th* feature domain).
  - If $L(D_i, D_j) = (|C_{ij}| < \epsilon)$, where $C_{ij} = \rho(D_i, D_j)$ and $\epsilon > 0$,
  - Then, $P(M_{ij}) > P(M_i)$
  Where
    - $M_{ij}$ is a model that combines $D_i$, $D_j$, and $M_i$ using only $D_i$.
    - $L(D_i, D_j)$: linear independence between $D_i$, and $D_j$
    - $C_{ij}$: Pearson correlation between $D_i$, and $D_j$
    - $\rho(D_i, D_j)$: Pearson's correlation function.
    - $\epsilon$: small positive threshold
    - $P(M_{ij})$: Performance metric of multimodal model $M_{ij}$
    - $P(M_i)$: Performance metric of multimodal model $M_i$
    - $|C_{ij}|$: Absolute value of $C_{ij}$

- Premise 2 (Linear Dependence and Interference)
  - If $H(D_i, D_j) = (|C_{ij}| > \delta)$, or $N(D_i, D_j) = (|C_{ij}| < -\gamma)$, where $C_{ij} = \rho(D_i, D_j)$ and $\delta, \gamma > 0$
  - Then, $P(M_{ij}) < P(M_i)$, or $P(M_{ijk}) < P(M_{ij})$
  - Where
    - $M_{ijk}$ is the model combining $D_i$, $D_j$, $D_k$
    - $H(D_i, D_j)$: Linear Dependence between $D_i$, $D_j$
    - $N(D_i, D_j)$: Negative Linear Dependence between $D_i$ and $D_j$
    - $\delta$: A positive threshold to define linear dependence
    - $\gamma$: A positive threshold to define negative linear independence
    - $P(M_{ijk})$: Performance metric of multimodal model $M_{ijk}$

- Premise 3 (Statistical Dependence and Information Overlap)
  - If $S(D_i, D_j) = (Mi_{ij} > \tau)$, where $Mi_{ij} = I((D_i; D_j)$ and $\tau > 0$, the model complexity increases without an appreciable increase in performance $P(M)$.
  - where
    - $S(D_i, D_j)$: Statistical dependence between $D_i$, and $D_j$
    - $Mi_{ij}$: Mutual information between $D_i$, and $D_j$
    - $I(D_i; D_j)$: mutual information function
    - $\tau$: Positive threshold for statistical dependence.

b) Experimental Evidence (Formalized Observation)
- Feature Domains: $D_1$(1D-CNN), $D_2$(2D-CNN), $D_3$(Transformer). ($D_i$: Represents the *i-th* feature domain)
- Observed Linear Relationships.
  - $L(D_1, D_2)$
  - $N(D_1, D_3)$
- Observed Statistical Dependence.
  - $S(D_i, D_j)$ observed for all $i,j$
- Observed Performance.
  - $P(M_{12}) > P(M_1)$
  - $P(M_{12}) > P(M_2)$
  - $P(M_{123}) < P(M_{12})$

c) Mathematical Reasoning
- Alignment of Theory and Experiment
  - Premise 1 explained $P(M_{12}) > P(M_1)$, $P(M_{12}) > P(M_2)$.
  - Premise 2 explained $P(M_{123}) < P(M_{12})$
  - Premise 3 explained the overall influence of shared information
- Mathematical Explanation of Performance Variation
  - $P(M)$ is a function of both correlations C and mutual information MI.
  $$P(M) = f(C, MI) \quad (1)$$
  - $L(D_i, D_j)$ minimizes redundancy but maximizes linear information gain.
  $$L(D_i, D_j) = \min_L [w_i . R(D_i, D_j) - w_i . G(D_i, D_j)] \quad (2)$$

- $H(D_i, D_j)$ maximizes redundancy but minimizes linear information gain.

$$H(D_i, D_j) = \min_{L}[w_i . G(D_i, D_j) - w_i . R(D_i, D_j)] \quad (3)$$

- $N(D_i, D_j)$ introduces conflicting signals, minimizing information gain.

$$N(D_i, D_j) = \min_{N} G(D_i, D_j) \quad (4)$$

- Where
  - R, G and w are redundancy, gain and weight respectively.

E. Theory: Complementary Feature Domains for Optimal ECG Multimodal Deep Learning Performance

- Empirical results revealed a counterintuitive trend: increasing the number of feature domains did not uniformly improve model performance; in fact, it sometimes led to degradation. Specifically, Hybrid 1, which combined 1D-CNN and 2D-CNN features, achieved an accuracy of 96%. However, the addition of a 1D-CNN transformer-based feature in Hybrid 2 resulted in a performance drop to 94%. This suggests that the Transformer features did not contribute complementary information. This decline was consistent across other evaluation metrics, including precision, recall, and F1 score, as detailed in Figure 19. This finding directly challenges the conventional wisdom that 'more is better' in terms of feature integration.
- The correlation, mutual information, Bayesian confidence interval, and bootstrapping analyses all agree with the statistical significance of the empirical results.
- Ablation studies agree with empirical findings and statistical significance tests: the removal of redundant features from an ECG multimodal model improves performance. Hybrid 1 outperformed Hybrid 2.
- The Scientific Reasoning Framework validates the central theory: feature domain complementarity is essential for improving ECG Multimodal Deep Learning Models. Redundancy adds complexity without any performance gain. The mathematical framework supports empirical findings, statistical significance tests, and ablation studies.
- Therefore, we do not find any evidence against the hypothesis: *While complementary feature domains enhance classification performance in hybrid multimodal deep learning models, incorporating redundant or non-discriminative domains increases complexity and suboptimal performance.*
- Thus, the performance of multimodal deep-learning ECG architecture depends on the complementarity of its feature domains.

Based on the outcomes of rigorous experimentation, statistical validation, ablation studies, and assertion by the scientific reasoning framework, the theory of **Complementary Feature Domains for Optimal ECG Multimodal Deep Learning Performance** is postulated. The novelty of this theory has the following significance:

- ✓ The rigorous logical, statistical and mathematical analysis to quantify feature domains' complementarity and redundancy while explaining the detrimental effects of conflicting linear patterns in the context of ECG multimodal deep learning.
- ✓ The approach provides a new perspective, departing from the conventional 'more is better' methodology. It offers concrete guidelines for designing optimal architecture in ECG multimodal deep learning, moving beyond reliance on purely heuristic feature selection techniques.
- ✓ The correctness of the hypothesis provides critical insights into optimizing hybrid multimodal ECG deep learning models, stressing the importance of balancing feature diversity with computational efficiency.
- ✓ Although the concept of complementarity has been used in various scientific disciplines, this study demonstrated its relevance in the context of ECG multimodal deep learning of time, frequency and time-frequency domains.

**Theory: Complementary Feature Domains for Optimal ECG Multimodal Deep Learning Performance** - *The complementarity of feature domains in a hybrid ECG multimodal deep learning model determines performance rather than the number of domains. Adding a redundant domain leads to plateaued or decreased model performance.*

Let:

- $D_1$ represent 1D-CNN features (time domain) extracted from ECG signals.

- $D_2$ represent the 2D-CNN features (time-frequency domain) extracted from ECG signals.
- $D_3$ represent Transformer features (frequency domain) extracted from ECG signals.
- $P(M)$ denotes the classification performance of a Hybrid Multimodal Deep Learning Model for ECG analysis.
- $I(D_i, D_j)$ denotes the mutual information between $D_i$ and $D_j$
- $\tau$ be a pre-defined threshold for mutual information

The proposed *Complementary Feature Domains for Optimal ECG Multimodal Deep Learning Performance theory* establishes the following principles:

1. **Claim 1: Complementary ECG Domains enhances Performance**

- If the two feature domains $D_i$ and $D_j$ derived from the distinct physiological aspects of ECG signals have non-overlapping information (i.e., minimal information sharing):

$$I(D_i, D_j) < \tau \text{ (i.e., low mutual information)}$$

Then, adding $D_i$ to a model that has $D_j$ will increase performance $P(M)$ in ECG classification.

- ✓ The empirical analysis showed a low correlation and low mutual information ($I(D_1; D_2) < \tau$) between the 1D-CNN ($D_1$) and 2D-CNN($D_2$) features. This represents a distinct physiological aspect of cardiac activity.
- ✓ In set theory terms, $D_i \cap D_j \approx \emptyset$ (i.e., the two feature domains intersect minimally, leading to broader information in the ECG analysis when merged).
- ✓ The $P(M_{12})$ performance of the hybrid model was higher than that of $P(M_1)$ or $P(M_2)$. This demonstrates the complementarity of $M_1$ and $M_2$.
- ✓ P(Hybrid 1) > P(1D-CNN) or P(2D-CNN).

2. **Redundancy plateaus or decreases performance**

- If the two feature domains $D_i$ and $D_j$ derived from the distinct physiological aspects of the ECG signals have overlapping information (i.e., non-negligible information sharing):

$$I(D_i, D_j) > \tau \text{ (i.e., high mutual information).}$$

Then, adding $D_i$ to a model that has $D_j$ will reduce or plateau performance $P(M)$ in ECG classification.

- ✓ In the empirical analysis, ($I(D_3; D_1) > \tau$) or ($I(D_3; D_2) > \tau$) between the transformer features ($D_3$) and 1D-CNN features ($D_1$) or 2D-CNN features ($D_2$) features.
- ✓ In set theory, $D_3 \subseteq D_2$ or $D_3 \subseteq D_1$. This results in an increase in the model complexity without a performance improvement.
- ✓ This suggests that the transformer features, which focus on the frequency domain, are redundant when merging with the 1D-CNN (time) and 2D-CNN (time-frequency).
- ✓ As predicted, performance $P(M_{123}) < P(M_{12})$. Empirical analysis showed that the transformer features contributed a conflicting linear pattern to the model.
- ✓ P(Hybrid 2) < P(Hybrid 1)

## 3. DISCUSSION

The complementarity or redundancy of the feature domains has proven to be a significant factor when considering the fusion of two or more domain features in an ECG multimodal deep learning architecture. Empirical findings corroborated statistical analysis, ablation studies, and scientific reasoning, clearly showing that adding more domain features does not necessarily translate to performance. While a complementary domain feature can boost performance, a redundant one can reduce or plateau it. Understanding the complementarity of the feature domains in multimodal deep learning models for ECG classification will provide a framework for designing efficient architectures. Fusing multiple domain features without incurring unnecessary complexity and overhead improves model performance. Consequently, this study will help advance the real-world applications of deep learning.

a) Domain Complementarity as a Design Principle for Multimodal Deep Learning

Figure 20 presents a scenario to demonstrate the complementarity of two distinct domains: the time domain (extracted via the 1D-CNN) and the time-frequency domain (extracted via the 2D-CNN). As shown in the figure, they captured the independent aspects of the ECG signal. Their fusion leads to improved performance and generalization.

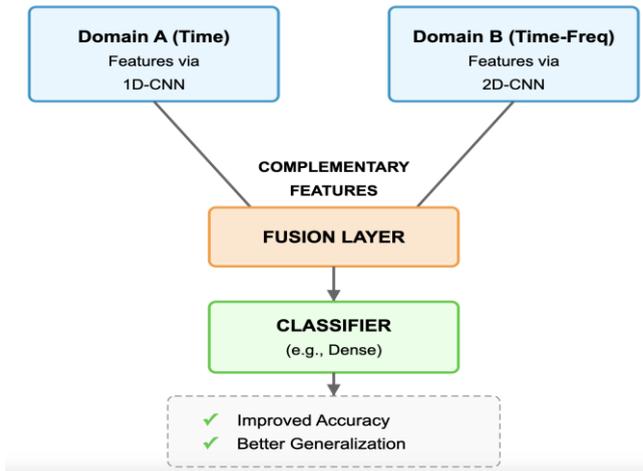

*Figure 20. Complementary Domains – Informative Fusion*

The complementarity between Domains A and B enables the classifier to generalize more effectively across unseen data, thereby improving the overall model performance. As demonstrated in this study, this fusion strategy achieves the highest classification accuracy (96%).

Figure 21 shows the case of redundancy between domains A and C. As shown, their fusion led to reduced accuracy, producing an overly complex model that generalized poorly (overfitting). This was empirically verified in this study; features from Hybrid 1 were fused with those from the transformer model. As shown in the results, the added feature domain did not improve the performance of Hybrid 1; rather, it reduced it (from 96% to 94%).

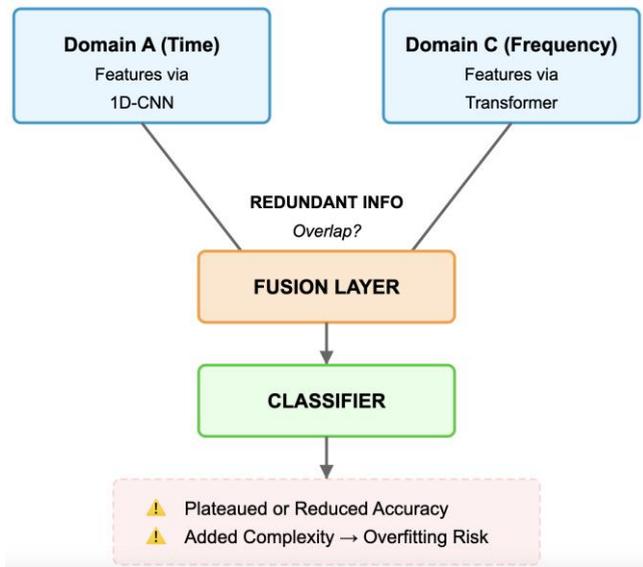

*Figure 21. Redundant Domains – No Performance Gain*

The scenario depicted in the framework in Figure 22 reinforces the centrality of the proposed theory, "Complementary Feature Domains for Optimal ECG Multimodal Deep Learning Performance Theory." As shown in the diagram, complementarity tests of domains should be performed before fusion. This study has shown that optimal performance in an ECG multimodal deep learning model is not solely a function of the number of fused domains but also of complementarity.

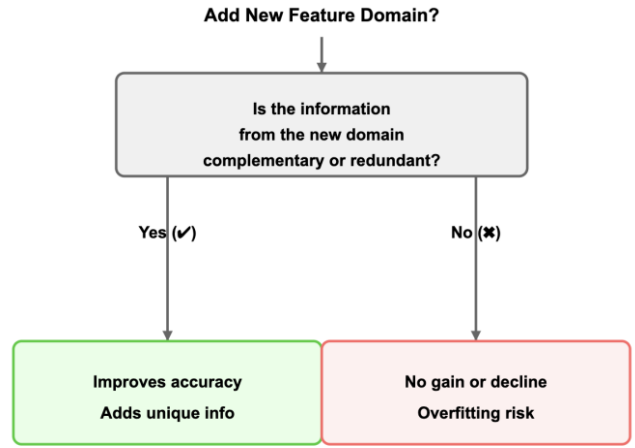

*Figure 22. Decision Flow Diagram*

b) Benchmarking Against Prior Multimodal ECG Fusion Strategies

While state-of-the-art classifiers seek to optimize disease detection accuracy [60, 61, 62], this study investigates the fundamental role of feature domain complementarity in multimodal architectures. The study aims to establish a theoretical framework for designing a multimodal ECG deep learning architecture. Table 18 contrasts this framework with traditional feature selection approaches that are often embedded in model pipelines. Implementation steps for the proposed theory using pseudocode are presented in Algorithm 4. A comprehensive mapping of the theory to the algorithm, along with a time complexity analysis, is presented in Tables 19 and 20, respectively.

*Table 18. Key Differences Between the Proposed Complementarity Feature Domain Theory and Traditional Feature Selection*

| Aspect | Complementary Feature Domain Theory (Proposed) | Traditional Feature Selection |
|---|---|---|
| **Goal** | Select **domain pairs** whose features are diverse but complementary, improving model generalization. | Select **individual features** with the highest relevance or statistical importance (e.g., via ANOVA, mutual information, or recursive elimination) |
| **Unit of Selection** | Entire **domain-specific feature vectors** (e.g., Time, Frequency, Time-Frequency) | **Scalar features** or small feature subsets (e.g., peak amplitude, variance) |
| **Selection Criterion** | Based on **pairwise domain complementarity**:<br>✓ Low mutual information<br>✓ High orthogonality | Based on **feature relevance** to label:<br>✓ High variance<br>✓ Strong correlation with class |
| **Representation Focus** | Emphasizes **cross-domain semantic diversity** | Emphasizes **feature-class statistical dependency** |
| **Fusion Objective** | Selects **optimal domain pairs** (e.g., F_Time and F_TF) that minimize redundancy and maximize representation coverage | Often combines top-k features, possibly from a single domain. |
| **Model Design Impact** | Leads to **modular encoders** per domain and **fusion modules** (e.g., cross-attention) | Leads to **single-branch** models fed by a filtered vector of features |
| **Interpretability** | Supports **domain-level interpretability** | Yields **fine-grained interpretability** but may lack high-level physiological meaning |
| **Used in Multimodal Context?** | ✓ Yes — designed for **multimodal learning** | X No — designed primarily for **unimodal tabular or signal features** |

*Complementarity Theory moves beyond traditional feature selection by prioritizing the diversity between entire feature sets from different domains, rather than focusing on the salience of individual features. It is especially useful in multimodal biosignal contexts like EEG or ECG where different domains encode fundamentally different physiological information.*

*Algorithm 4. Theory-Guided ECG Multimodal Deep Learning via Complementary Feature Domains*

**Theory:** *The complementarity of feature domains in a hybrid ECG multimodal deep learning model determines performance rather than the number of domain features. Adding a redundant domain leads to plateaued or decreased model performance.*

**Inputs:**
ECG_signal  // Raw ECG input signal
D = {$D_1, D_2, ..., D_n$}  // (e.g., Time, Frequency, Time-Frequency)
F_extractor  // (e.g., 1D-CNN, 2D-CNN, Transformer)
$\tau$_MI, $\tau$_ortho  // Thresholds for Mutual Information and Orthogonality respectively
$\lambda_1, \lambda_2$  // Weights for complementarity-aware loss terms

**Output:**
ŷ  Predicted class label (e.g., CVD category)

**Steps:**

1: Preprocess ECG_signal:
  ▷ Apply denoising, baseline correction, and segmentation

2: Extract Domain-Specific Features:
  For each $D_i \in D$:
    $X_i \leftarrow D_i$(ECG_signal)
    $F_i \leftarrow$ F_extractor$_i(X_i)$

3: Assess Pairwise Complementarity
  ComplementarySet $\leftarrow \emptyset$
  For each pair ($F_i, F_j$), $i \neq j$:
    $MI_{ij} \leftarrow$ MutualInformation($F_i, F_j$)
    $Ortho_{ij} \leftarrow \|F_i^T \cdot F_j\|_2$  // Frobenius norm
    If $MI_{ij}$ < T_MI and $Ortho_{ij}$ < T_Ortho:
      Add ($F_i, F_j$) to ComplementarySet

4: Select optimal domain pair(s):
  ▷ Choose the pair (F_opt_i, F_opt_j) maximal complementarity:
  - Low mutual information (minimal redundancy)
  - High orthogonality (maximal synergy)
  (F_opt_i, F_opt_j) ← select_best_pair(ComplementarySet)

  ▷ Exclude any third domain $F_k$ violating complementarity to the selected pair

5: Fuse selected domain features:
  F_fused ← FusionModule(F_opt_i, F_opt_j)
  ▷ Fusion types: cross-attention, intermediate, or weighted average

6: Predict output:
  ŷ ← Classifier(F_fused)

7: Compute complementarity-aware loss:
  ▷ This is the core novelty: enforce domain synergy through learned optimization
  ▷ Unlike standard regularization (e.g., L2), this Algorithm operates on relationships between domain features

  L_class ← CrossEntropyLoss(ŷ, y_true)  Task loss
  L_MI ← MutualInformation(F_opt_i, F_opt_j)  Penalize redundancy
  L_Ortho ← $\|$F_opt_i$^T$ · F_opt_j$\|_2$  Reward orthogonality

  L_complementary ← L_class + $\lambda_1$ · L_MI + $\lambda_2$ · L_Ortho   Final total loss

8: Backpropagation:
  Update model parameters using gradient $\nabla$L_complementary

9: Evaluation and validation:
  ▷ Compute classification metrics (Accuracy, F1-score, etc.)
  ▷ Visualize saliency maps for interpretability
  ▷ Perform ablation to confirm performance drops with redundant domains

**Return:**
ŷ

*Table 19. Theory-to-Algorithm Mapping*

| Theoretical Component | Algorithm Implementation |
|---|---|
| Performance depends on feature **domain** complementarity, not quantity | Steps 3–4 use Mutual Information (MI) and Orthogonality to select only complementary pairs |
| Redundant domains cause degradation. | Step 4 excludes domains with high MI (redundancy) or low orthogonality |
| Fusion of complementary feature domains | Step 5 fuses only validated, complementary feature domains |

| Empirical validation of theory | Step 9 includes evaluation of both classification metrics and complementarity metrics |
| Penalize non-complementary features | Step 7 adds loss terms for mutual information and orthogonality violation |

**Assumptions for Complexity Notation**
- d - Number of domain transformations (e.g., time, frequency, time-frequency)
- n - Number of ECG samples (instances in the dataset)
- $T_i$ - Time complexity of the feature extractor for domain $D_i$
- m - Dimensionality of the feature vector output from each extractor
- Assume fixed-size segments and model depth for CNN/Transformer modules

*Table 20. Time Complexity Analysis*

| Step | Description | Time Complexity (Per Epoch/Iteration) |
|---|---|---|
| 1 | Preprocess ECG signal | $O(n)$ |
| 2 | Extract candidate feature vectors | $O(n \cdot d \cdot m)$ |
| 3 | Evaluate pairwise domain complementarity | $O(n^2 \cdot d^2)$ |
| 4 | Select optimal feature pair(s) | $O(d^2)$ |
| 5 | Fuse selected features | $O(n \cdot T_{fusion}(m))$ |
| 6 | Predict | $O(n \cdot T_{classifier}(m))$ |
| 7 | Compute loss | $O(n \cdot (T_{MI}(m) + m))$ |
| 8 | Backpropagation | $O(n \cdot [T_{preprocess} + d \cdot T_F + d^2 \cdot (T_{MI}(m) + m) + T_{fusion}(m) + T_{classifier}(m)])$ |
| 9 | Evaluate | $O(n + T_{saliency})$ |

The total time complexity of the algorithm per training epoch is dominated by the backpropagation step.

**Interpretation**
- The total complexity is linear with respect to the number of samples n.
- Complexity increases quadratically with the number of domains d due to pairwise complementarity checks.
- Feature vector dimensionality m influences the cost of MI and orthogonality evaluation, but is typically fixed.
- Efficient when d is small (e.g., time, frequency, time-frequency).

**Bottleneck**
- The primary bottleneck is the complementarity evaluation (Step 3), especially for Mutual Information (MI) and orthogonality, which scale as $O(d^2 \cdot m^2)$.
- MI estimation can be expensive if non-parametric (e.g., k-NN). Faster approximations, such as MINE or variational MI, are recommended for improved scalability.

c) Comparative Evaluation Demonstrates Domain-Level Hybrid Fusion Outperforms Traditional Multimodal Strategies

Recent research in cardiovascular signal classification has increasingly favored multimodal data fusion strategies, integrating diverse biosignals such as electrocardiography (ECG), phonocardiograms (PCG), and hemodynamic waveforms (e.g., BP, ART, PAP). These approaches often attempt to compensate for modality-specific noise or variability by incorporating complementary physiological measurements. However, this comes at the cost of increased hardware requirements, synchronization complexity, and reduced interpretability. In contrast, this work investigates whether complementary representations derived from intra-modal signal transformations of ECG alone can match or surpass the performance of inter-modal multimodal systems.

The proposed Hybrid 1 architecture, which employs complementary fusion of time-domain and time-frequency domain features (via 1D-CNN and 2D-CNN pipelines, respectively), was benchmarked against two recent multimodal approaches: (1) Khavas & Mohammadzadeh Asl [63], which integrates ECG with BP, ART, and PAP signals using a late rule-based fusion strategy; and (2) Coimbra et al. [64], which combines ECG and PCG via early-fusion in a deep learning framework with explainable AI components.

As summarized in Table 21, the proposed Hybrid 1 model achieved an accuracy of 96.0% and an F1-score of 95.7%, significantly outperforming Khavas et al. (accuracy: 93.1%, F1-score: 95.6%) and Coimbra et al. (F1-score: 86.0%, with a reported AUC 0.81 for which a direct accuracy comparison was not feasible). This superior performance is achieved using only ECG-derived domain features, eliminating the need for multi-sensor configurations while maintaining a high degree of physiological interpretability. These findings empirically support the hypothesis that feature-level complementarity within a single biosignal, when extracted and fused via domain-aware architectures, can yield classification performance that exceeds more complex multimodal setups.

Furthermore, the use of intermediate fusion facilitates the combination of deep temporal and spectro-temporal representations, creating a unified embedding space with greater discriminative capacity. Unlike ensemble-based or late-decision fusion schemes, the proposed model is fully end-to-end trainable, lightweight, and readily adaptable to real-world ECG acquisition pipelines such as wearable devices or telehealth systems. By relying exclusively on ECG and exploiting internal domain complementarity, the proposed method offers a more scalable, interpretable, and clinically deployable solution for automated cardiovascular risk stratification.

*Table 21.Comparative Benchmarking with Recent Multimodal Fusion Models*

| Study | Year | Modalities | Fusion Strategy | Accuracy (%) | F1-Score (%) |
|---|---|---|---|---|---|
| Khavas & Mohammadzadeh Asl [63] | 2018 | ECG + BP + ART + PAP | Late (Rule-Based) | 93.1 | 95.6 |
| Coimbra et al. [64] | 2024 | ECG + PCG | Early | ~81.0 (AUC) | 86.0 |
| **Proposed Hybrid 1** | 2025 | ECG Time + Time-Frequency | Intermediate | **96.0** | **95.7** |
| **Hybrid 2** | 2025 | ECG Time + Time-Freq + Frequency | Intermediate | 94.3 | 93.7 |

d) Explainability of Hybrid 1 (1D-CNN + 2D-CNN )

Figure 23 displays the saliency (red) map for the hybrid 1D-CNN + 2D-CNN model applied to the four classes, correctly classified ECG signals: one per class (*Class 0: MI, Class 1: History of MI, Class 2: Abnormal Heartbeat, Class 3: Normal*). The figure presents class-wise saliency overlays for the Hybrid1 model, illustrating how model attention aligns with critical ECG waveform components. The black line represents the raw ECG signal, while the red line denotes the corresponding saliency signal, indicating regions that contribute most strongly to the model's prediction. For each class, the red saliency trace highlights regions deemed most informative by the model. Notably, consistent alignment with the QRS complex and T-wave is observed across classes, indicating physiologically meaningful feature attribution.

Complementing the saliency map is Figure 24 (bottom), which displays a radar plot of four quantitative XAI metrics: Peak Similarity, Temporal Consistency, Spectral Similarity, and Stability Index, normalized by metric. Class 2 consistently achieves high interpretability scores across all dimensions, while Class 0 lags in temporal and spectral alignment. These quantitative insights validate the visual saliency patterns, reinforcing that Hybrid1 captures class-discriminative and clinically relevant ECG features with interpretive consistency across modalities. The results demonstrate that the model learns physiologically relevant decision patterns distinct to each class. ECG Deep Learning Multimodal Explainability has been well documented in earlier work [65].

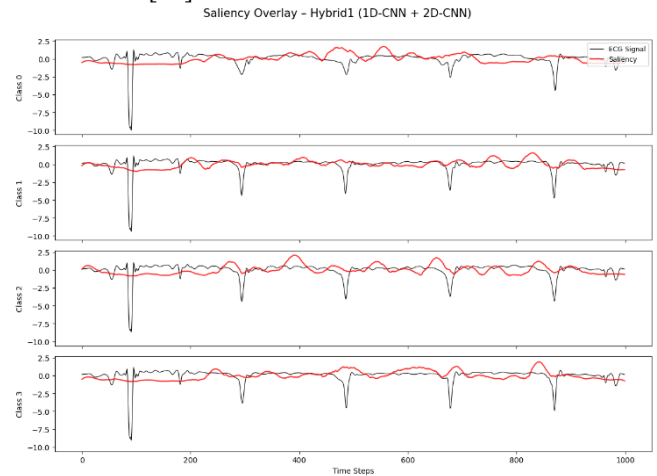

**Figure 23. Saliency Overlay for Hybrid1 Model (1D-CNN + 2D-CNN).** *Overlay of the original ECG signal (black) and the corresponding saliency signal (red) for the four CVD classes. Saliency values are computed using gradient-based attribution methods and smoothed with a Gaussian filter for interpretability. Notably, saliency peaks consistently align with physiologically relevant regions (e.g., QRS complex), demonstrating the model's focus on clinically meaningful patterns. This alignment supports the interpretability and trustworthiness of the Hybrid1 architecture, validating the complementary contribution of time and time-frequency domains.*

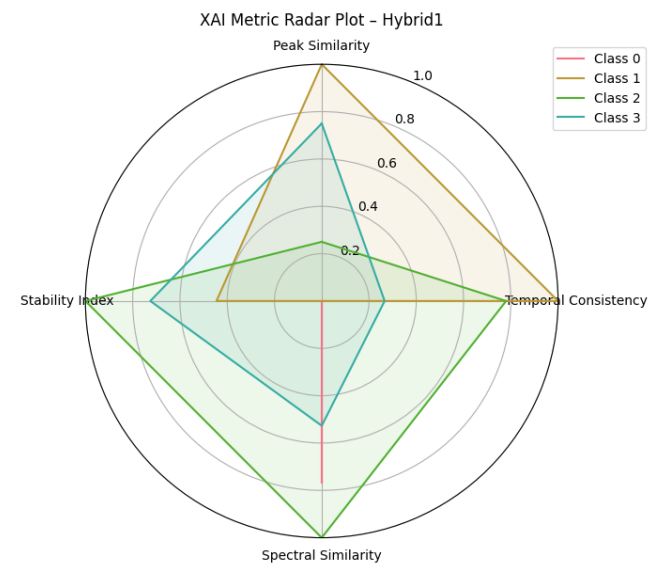

**Figure 24. Explainability Metric Radar Plot for Hybrid1 (1D-CNN + 2D-CNN).** *Quantitative interpretability assessment of the Hybrid1 model across the four classes using four saliency-based explainability metrics: Peak Similarity, Temporal Consistency, Spectral Similarity, and Stability Index*

e) Overview of Multimodal Fusion Strategies

Beyond intermediate fusion, several other fusion strategies have been employed in multimodal deep learning.
- **Late fusion** operates at the decision level by combining the outputs (e.g., softmax probabilities or class scores) of independently trained unimodal models using weighted averaging, majority voting, or meta-learners [66].
- **Early fusion** involves concatenating raw or low-level features before learning begins, allowing joint optimization but often suffering from heterogeneity in feature distributions and scale mismatches [67].
- **Attention-based fusion** mechanisms, such as self-attention or cross-attention, have been used to weigh the contributions of each modality or feature dimension dynamically. These mechanisms can capture intricate interdependencies across modalities, but typically require more parameters and data [68].
- **The intermediate fusion**, as adopted in this study, strikes a balance between feature expressiveness and model complexity by integrating learned feature representations from each domain prior to classification.

*f)* Application of the Proposed Theory

While the current study focuses on ECG signal classification, the proposed framework, centered on the concept of complementary feature domains and intermediate feature-level fusion, is modality-agnostic and can be adapted to other biomedical and time-series domains.

I. Complementarity Theory Application in Electroencephalography (EEG)-based Seizure Detection

The proposed theory applies to a domain-complementarity-based pipeline for seizure detection using EEG signals. The framework begins with raw multi-channel EEG input, which undergoes standard preprocessing steps, including filtering, artifact removal, and segmentation. The preprocessed signal is then broken down into three separate domains: Time, Frequency, and Time-Frequency.

- **Time-domain features** are extracted via a 1D-CNN applied to raw EEG waveforms, capturing waveform morphology and spike events.
- **Time-frequency features** are computed via CWT or STFT, then processed using a 2D-CNN to capture spectral bursts and phase shifts.
- **Frequency-domain features** are derived by applying a Fast Fourier Transform (FFT) followed by a Transformer, targeting oscillatory bands.

II. Complementarity Theory Application in Human Activity Recognition

The proposed theory also applies to a multimodal deep learning architecture designed for human activity recognition (HAR) that utilizes raw accelerometer signals. The principle of feature domain complementarity ensures that combining time-domain and frequency-domain features results in improved discriminative ability and model efficiency. The model takes Raw Acceleration (e.g., 3-axis X, Y, Z Accelerometer Data) and processes it through Algorithm 4 to produce a learned representation of activity categories as output.

The above two examples illustrate the broader utility of the architectural approach in multimodal or multi-domain settings, supporting the theory that **fusion success is governed more by feature domain complementarity than by the number of fused representations**. The robustness of the design is crucial for the success of multimodal deep learning in improving classification performance.

F. FUTURE WORK
- We will explore different concatenation techniques such as attention-based fusion, graph neural networks (GNNs), and autoencoders
- Integrating the transformer feature domain into Hybrid 1 reduced the performance from 96% to 94%. In the future, we will optimize the transformer-based architectures.
- We will integrate Explainable Artificial Intelligence (XAI) techniques for model interpretation. This enhances our study of the complementarity of the feature domains.
- This study used an ECG signal dataset, and we hope to use a more diverse dataset in the future.

4. CONCLUSION

This study examined how various combinations of extracted features from different signal processing domains (time, frequency, and time-frequency) and learned deep features from diverse learning algorithms (1D-CNN, 2D-CNN, and transformers) affect the performance of a multimodal deep learning model. An optimal balance that maximizes performance without incurring unnecessary complexity is also highlighted. An ECG signal dataset from the public domain is used in this study. The dataset was loaded into Google Colab, and scripting was performed using Python in the Keras environment. Algorithm descriptions and implementation guidelines were provided in tabular form to ensure the reproducibility of the experiment.

An empirical investigation and statistical analysis demonstrated the impact of feature complementarity on the performance of a hybrid multimodal deep learning model. Ablation studies and formal scientific reasoning have confirmed the reliability of these findings. Therefore, ECG multimodal deep learning models should be trained using domain features that are both necessary and sufficient for optimal performance. Consequently, the proposed theory, "**Complementary Feature Domains for Optimal ECG Multimodal Deep Learning Performance**," aligns with Occam's razor principle of parsimony and plurality [69, 70], Newton's first rule of scientific reasoning [71, 72], and the Minimum Description Length (MDL) principle [73, 74].

Thus, the performance of a hybrid ECG multimodal deep learning model is not solely determined by the diversity and number of concatenated feature domains, but by their complementarity. The optimal classification results depend on selecting non-redundant, complementary feature domains and maintaining simplicity in the model's architectural complexity.

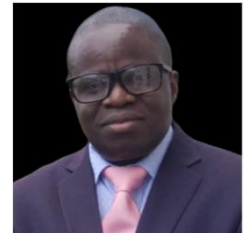

Timothy Oladunni received the master's and Ph.D. degrees in computer science from Bowie State University, MD, USA, in 2013 and 2017, respectively. He is currently an Assistant Professor of Computer Science at Morgan State University, MD, USA. He was a Visiting Assistant Professor/Faculty Research Fellow in the Computer Science Department at Yale University, CT, USA.

Timothy is a distinguished computer scientist, professor, and machine learning researcher specializing in biomedical signal processing, natural language processing, deep learning, and multimodal AI architecture. With a background in electrical engineering, he has dedicated his research to advancing ECG signal analysis, natural language processing, and pattern recognition. His recent work has focused on multimodal deep-learning architecture, particularly the trade-off between model complexity and performance in biomedical signal classification. By integrating the time, frequency, and time-frequency domain features, he explores novel ways to optimize CNN-transformer-based models for ECG analysis, ensuring robust and generalizable AI-driven diagnostic systems.

As a professor, Timothy is passionate about mentoring the next generation of data scientists and AI researchers.

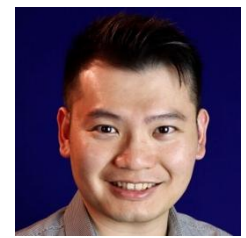

Alex Wong is an Assistant Professor in the department of Computer Science at Yale University. Prior to joining Yale, he was an Adjunct Professor at Loyola Marymount University (LMU) from 2018 to 2020. He received his Ph.D. in Computer Science from the University of California, Los Angeles (UCLA) in 2019 and was previously a post-doctoral research scholar at UCLA from 2020 to 2022. His research lies in the intersection of machine learning, computer vision, and robotics and largely focuses on multimodal 3D reconstruction, robust vision under adverse conditions, and unsupervised learning.